\documentclass{ieeeaccess}
\usepackage{cite}
\usepackage{amsmath,amssymb,amsfonts}
\usepackage{algorithm,algorithmic}
\usepackage{graphicx}
\usepackage{multirow}
\usepackage{textcomp}
\usepackage{subfigure,caption}
\usepackage{booktabs}
\usepackage{threeparttable}
\usepackage{grffile}
\usepackage[super]{nth}
\usepackage[T1]{fontenc} 
\newcommand*{\permcomb}[4][0mu]{{{}^{#3}\mkern#1#2_{#4}}}

\newcommand*{\comb}[1][-1mu]{\permcomb[#1]{C}}
\newcommand{\degree}{\ensuremath{^\circ}}

\def\BibTeX{{\rm B\kern-.05em{\sc i\kern-.025em b}\kern-.08em
		T\kern-.1667em\lower.7ex\hbox{E}\kern-.125emX}}

\begin{document}
	\history{Date of publication xxxx 00, 0000, date of current version xxxx 00, 0000.}
	\doi{10.1109/ACCESS.2017.DOI}
	
	\title{A non-invertible cancelable fingerprint template generation based on ridge feature transformation}
	\author{\uppercase{Rudresh Dwivedi}\authorrefmark{1}, 
		\uppercase{Somnath Dey}\authorrefmark{2}, \IEEEmembership{Member, IEEE}}
	\address[1]{Discipline of Computer Science \& Engineering,  \\
		Indian Institute of Technology Indore, Indore, India (e-mail: phd1301201006@iiti.ac.in)}
	\address[2]{Discipline of Computer Science \& Engineering,  \\
		Indian Institute of Technology Indore, Indore, India (e-mail: somnathd@iiti.ac.in))}
	
	\tfootnote{The authors are thankful to SERB (ECR/2017/000027), Department of science \& Technology, Govt. of India for providing financial support to carry out this research work.}
	
	\markboth
	{R. Dwivedi \headeretal: A non-invertible cancelable fingerprint template generation based on ridge feature transformation}
	{R. Dwivedi \headeretal: A non-invertible cancelable fingerprint template generation based on ridge feature transformation}
	
	\corresp{Corresponding author: Rudresh Dwivedi (e-mail: phd1301201006@ iiti.ac.in).}
	
	\begin{abstract}
		In a biometric verification system, leakage of biometric data leads to permanent identity loss since original biometric data is inherently linked to a user. Further, various types of attacks on a biometric system may reveal the original template and utility in other applications. To address these security and privacy concerns cancelable biometric has been introduced. Cancelable biometric constructs a protected template from the original biometric template using transformation functions and performs the comparison between templates in the transformed domain. Recent approaches towards cancelable fingerprint generation either rely on aligning minutiae points with respect to singular points (core/delta) or utilize the absolute coordinate positions of minutiae points. In this paper, we propose a novel non-invertible ridge feature transformation method to protect the original fingerprint template information. The proposed method partitions the fingerprint region into a number of sectors with reference to each minutia point employing a ridge-based co-ordinate system. The nearest neighbor minutiae in each sector are identified, and ridge-based features are computed. Further, a cancelable template is generated by applying the Cantor pairing function followed by random projection. We have evaluated our method with FVC2002, FVC2004 and FVC2006 databases. It is evident from the experimental results that the proposed method outperforms existing methods in the literature. Moreover, the security analysis demonstrates that the proposed method fulfills the necessary requirements of non-invertibility, revocability, and diversity with a minor performance degradation caused due to cancelable transformation.
	\end{abstract}
	
	
	\titlepgskip=-15pt
	
	\maketitle
	
	\section{Introduction}
	\PARstart{T}{he} compromise of stored biometric template causes permanent identity theft of a user as biometric data is irreplaceable and irrevocable. There are various types of attacks and privacy concerns linked with sharing of biometric information across multiple applications \cite{crack,anneal}. Jain et al. \cite{bts} identified four levels of attacks in a biometric system. At the first level, the attacker presents a falsified biometric input to the sensor, and the sensor may not be able to differentiate between genuine and fake biometric inputs. Second, the attacker intercepts the communication link between the different modules to enter into the system. At the third level, the attacker intercepts the executable program of a module to get the desired output. Such attacks are called Trojan-Horse attacks. Last, the attacker replaces/ derives a spoof of the stored template which causes the security breach. Therefore, biometric template protection is necessary to address such security challenges. To provide biometric template protection against the afore-mentioned attacks, an idea of cancelable biometrics has been presented. The cancelable biometric scheme applies a transformation to derive a protected template which is used for verification instead of the original template of a user. The transformation relies on a non-invertible function such that it is hard to retrieve the original template even if the attacker knows the transformed template and transformation function. In a compromise situation, a new protected template can be derived by altering the parameter values of the transformation function. The transformation should fulfill the following requirements described by Breebaart et al. \cite{iso}:
	
	\begin{enumerate}
		\item Non-invertibility: It should be computationally hard to construct the original template from the transformed template. This prevents the recovery of the original biometric information by an imposter. 
		\item Diversity: Identical cancelable template should not be used in different applications to avoid cross-matching of the stored template. 
		\item Revocability: The transformation should be able to derive numerous protected templates from the same biometric input and there should be immediate revocation in case of compromise.
		\item Performance: The transformation should not exhibit significant performance degradation.
	\end{enumerate}
	
	Bolle et al. \cite{bolle} introduced four types of transformations for cancelable fingerprint template generation which utilize the core and delta points (singular points) of an input fingerprint. The biohashing based approaches proposed in \cite{biohash,hash2} combine a user-specific key onto the original biometric features to derive a protected template. BioHashing and its variants \cite{biohash, hash2}, are proved to be impractical if the unique seed is compromised. Ahmad et al. \cite{pairp} proposed a fingerprint template protection scheme that uses relative minutiae information in the polar coordinate system as described in \cite{tuple}. The proposed scheme underperforms on the FVC2002 DB3 database where low-quality fingerprint images exist. The performances of the techniques presented in \cite{tuple,cct,mlc,lee} get degraded if the user-specific key or token is compromised. Further, the limitation of the methods proposed in \cite{bolle,das,rlrd} lies with the accurate detection of the singular point (core or reference point). However, it is not possible to identify the singular points from fingerprint images of all users. Moreover, accurate detection of core point from an arch type or a poor quality fingerprint image is a challenging task. 
	
	Farooq et al. \cite{farooq} derived a binary representation of the different features computed from all possible triplets which require a large number of computations. Cappelli et al. \cite{mcc} introduced a state-of-the-art Minutiae Cylinder Code (MCC) based fingerprint representation which outperforms the most of the existing methods. However, the original MCC approach does not provide any protection mechanism for minutiae information. Later, Ferrara et al. \cite{pmcc} proposed the protected MCC (P-MCC) approach to secure the minutiae information. Further investigations reveal the irrevocability issue of P-MCC technique. In order to provide revocability, Ferrara et al. \cite{2pmcc} proposed a two-factor protected Minutiae Cylinder Code (2P-MCC) scheme which performs curtailed permutation onto cylinders in P-MCC using a secret key.
	
	Few approaches \cite{rlrd,sutcu} in literature utilize fixed-radius transformation. These approaches may cause performance degradation if the minutiae points are at the edge of the radius. Owing to noise or local distortion these minutiae could be considered inside the radius for the first sample and outside the radius for the second sample for the same fingerprint. Further, Ahmad et al. \cite{pairp} and Sutcu et al. \cite{sutcu} applied a transformation considering a threshold onto the number of minutiae point to derive the protected template. Certain methods \cite{lee,mlc} directly use the position and direction information of minutiae points to derive a protected fingerprint template. However, selection of invariant features from the minutiae points results in significant performance improvement over the original minutiae information. Earlier, we have proposed a coprime mapping based transformation to protect the original ridge features. In the prior work \cite{mymike}, a very less number of brute-force attempts are required yielding the approach invertible. Moreover, the method is susceptible to the Attack-via-record-multiplicity (ARM) attack. Also, the performance degrades in case of FVC2004 where the users are allowed to exaggerate deformations at the time of acquisition. In this work, we have proposed random projection based transformation to mitigate the limitations of our prior work.
	
	In summary, to address the limitations of the existing approaches depicted above, we propose a novel cancelable fingerprint template generation method based on ridge feature transformation. Ridge-based features are computed for the nearest neighbor structure drawn for each reference minutiae point. Next, the Cantor pairing function is applied to encode the ridge features, and the logarithm function is used to uniformly distribute the paired features. Finally, the random projection is utilized to derive a non-invertible protected template. In a nutshell, we highlight the contributions of our work:
	
    \begin{enumerate}
	\item We have proposed a novel cancelable fingerprint template design methodology where transformation is applied over ridge features to cope with rotation, translation and scale deformations in the input fingerprint image.
	\item The proposed transformation does not lean upon prior alignment with the singular point which is hard to discover in the poor quality or missing singular point fingerprint images.
	\item The ridge feature transformation is applied around each minutia to derive a feature matrix instead of fixed-radius transformation to overcome the boundary problems.
	\item The Cantor pairing function followed by random projection is utilized to generate non-invertible cancelable fingerprint template.
	\item The proposed scheme is analyzed with respect to the necessary criteria of cancelable template generation i.e. non-invertibility, revocability and diversity. Further, the method is also analyzed against different types of attacks such as Attack via Record Multiplicity (ARM), pre-image attack, cross-matching attack, distinguishing attack and annealing attack. The security analysis demonstrates that the proposed approach fulfills the desired criteria and is robust enough to sustain such attacks.
	\item The performance of the proposed method is evaluated with two different protocols (FVC and 1VS1) for all datasets (i.e. DB1, DB2, DB3 and DB4) of FVC2002, FVC2004 and FVC2006 databases. The experimental results demonstrate that our approach outperforms existing approaches.
\end{enumerate}
The rest of the paper is organized as follows. In Section II, we briefly summarize the existing methods related to the generation of cancelable fingerprint template. Section III describes the procedural steps of the proposed method. Section IV demonstrates experimental results as well as compares the proposed method with the existing cancelable template generation approaches. Section V provides the security analysis of our method. Conclusions and direction of future research are presented in Section VI.

\section{Related work}
In recent years, several methods have been introduced to generate the cancelable template in the literature. Ratha et al. \cite{ratha} first introduced the notion of cancelable biometric with three different types of transformations (Cartesian, polar and functional) to provide privacy and security to the original biometric information. The cartesian transformation method maps the fingerprint minutiae into cells of fixed size. The minutiae positions in the cells are permuted to derive the cancelable template. In polar transformation method, the minutiae positions are mapped onto a polar coordinate space. Further, the coordinate space is divided into sectors, and sector positions are shuffled based on a key to generate the cancelable template. Functional transformation method alters the minutiae positions and orientation based on a parametric Gaussian function. Quan et al. \cite{crack} proved that the functional transformation could be cracked if the parameters and transformed template are revealed. The rest two transformations cater with high Equal Error Rate (EER). Moreover, all these methods require core-point for alignment before the transformation, and the determination of core-point is not always feasible.

Boult et al. \cite{boult} proposed secure biotokens for fingerprint template protection. The method constructs a minutiae pair table which contains distance, relative orientation, and orientation of the line connecting two minutiae points. The features are then divided into quotient and modulus part. The quotient part is encrypted using RSA algorithm and modulus part is concatenated with the encrypted quotient to form a cluster. In verification stage, minutiae pair tables of the query and stored templates are traversed to construct clusters and to compute the comparison scores. Lee et al. \cite{lmi} proposed an alignment-free protected fingerprint template generation scheme using minutia orientation. They applied two different changing functions (i.e., positions and respective orientations) which are used to secure the minutiae information. The stored template can thus be regenerated and revoked by altering the input parameters of the changing functions in the situation of compromise. In BioHashing based approaches \cite{biohash,hash2}, the protected template is derived after discretizing the inner product of the biometric features with the projection matrix. Yang et al. \cite{byang} proposed a non-linear dynamic random projection scheme to increase the computational complexity against the inversion attack. Instead of conventional random projection utilized by BioHashing, the projection matrix is dynamically constructed based on an index vector which is constructed from the quantization of the biometric feature vector. Lee et al. \cite{lee} presented a method to derive cancelable fingerprint template based on 3-D array mapping. In this work, a minutia from the minutiae set is assigned as the reference, and remaining minutiae are aligned with respect to the reference minutiae. Then, the aligned minutiae points are mapped into a 3-D array, based on the positions (\textit{x}-\textit{y} co-ordinate) and orientations of the minutiae points. The cells in the 3-D array are marked as 1, which include minutiae points. The array is sequentially traversed to derive a bit-string. The derived bit-string is exploited to random permutation utilizing a user-specific PIN and minutiae type. Wang et al. \cite{ditom} proposed a template protection mechanism where many-to-one mapping is applied onto the pair-minutiae based bit-string evaluated using the method proposed by Lee et al. \cite{lee}. Then, the user-specific PIN is applied to the complex vector derived by discrete Fourier transform on bit-string. Wang et al. \cite{cct} proposed another method for cancelable fingerprint template design using circular convolution. The procedure adopted till bit-string generation is identical to it's earlier method \cite{ditom}. Then, a random sequence is derived by utilizing a user-specific PIN. Bit-string and random sequence are exploited to discrete Fourier transform and product of both DFT's are computed. The cancelable template is stored by applying inverse-DFT and removing the first $(p-1)$ points from the output. Das et al. \cite{das} constructed a graph structure based on the minimum possible distance from core/delta point to remaining minutiae points. Correspondence search algorithms \cite{das} are used for verification of query template. Liu et al. \cite{rlrd} proposed a template protection scheme which derives a protected fixed-length template viz., random local region descriptor (RLRD). In this scheme, a random reference point is selected initially. Next, Tico's sampling structure \cite{rlrd} is utilized to generate uniformly distributed sampling point structure around the random reference. The order of the sampling points is decided through a random seed. Finally, a protected template is derived as the angular width between the reference and sampling points. Further, gray-code encoding of sine and cosine of angular width is performed to generate a bit-string. Wong et al. \cite{mlce} proposed a multi-line code (MLC) for minutiae-based fingerprint template protection which is an extension of Wong et al. \cite{mlc} work. In this method, the minutia set is divided into angular partitions with respect to a straight line drawn at the reference minutiae. Next, few sample points with equal distance to each other are taken with uniform distribution on the straight line. Circles are constructed on each of the sample points, and minutiae points falling in each lower region (semi-circle) are counted. A binary string is derived considering 1 if the count is greater than 0, and 0 otherwise. The result is stored as the cancelable template. In their extended work \cite{mlce}, the mean distance from minutiae to the line is computed in each semicircle along the lines. The quantization is performed over the mean distances. The binary string is derived and permuted with a user-specific PIN to generate the cancelable template. Ahmad et al. \cite{pairp} proposed a cancelable fingerprint template design scheme using many-to-one sector mapping for relative minutiae information in the polar coordinate system as described in \cite{tuple}.

Farooq et al. \cite{farooq} proposed a triangular transformation which derives a binary representation of minutiae features. The method utilizes the minutiae triplet features: length of each side, the angle subtended between each side, each minutiae orientation in the triplet and the height of triplet. These features are quantized into 24 bits to derive a $2^{24}$-bit binary representation. Sutcu et al. \cite{sutcu} introduced a geometric design which represents minutiae information into a fixed length string. The method computes mean of minutiae \textit{x} and \textit{y} coordinate as centroid. Next, a circle is constructed centered at centroid and divided into an arc of equal angular width.  Then, a straight line is drawn in between each minutiae pair and intersection with the circumference is marked. The number of intersection mark is collected sequentially for each arc to derive the transformed template. Wang et al. \cite{blind} proposed a scheme which utilizes DFT for pair-minutiae bit string. Further, the complex sequence is fed to Finite-Impulse-Response (FIR) with a user-specific key to derive the protected template. The method performs optimally yet weak against attack via record multiplicity (ARM). In their future work, Wang et al. \cite{hadamard} proposed a partial Hadamard based transformation the to protect original pair minutiae bit-string. 

Cappelli et al. \cite{mcc} introduced a state-of-the-art Minutiae Cylinder Code (MCC) algorithm which frames a 3-D cylindrical structure around minutiae neighborhood considering each minutia as a reference. Each cylinder of height $2\pi$ and radius $r$ is tessellated into a number of cells. Each cell stores the minutiae position and orientation in the neighborhood of each minutia taken as a reference at a time. A cylinder which contains less valid information is discarded. Two cylinders are verifiable if direction difference between two minutiae is less than a certain value. MCC is a fixed radius local minutiae construct which provides notable recognition performance. However, high computation cost of cell construction for each cylinder is the drawback associated with this method. Ferrara et al. \cite{pmcc} then proved that few genuine minutiae points (approximately 25.4\%) could be correctly revealed by calculating likelihood between two cylinders. Later, a novel representation namely Protected-MCC (P-MCC) is proposed where a non-invertible transform has been applied onto MCC template incorporating binary-KL projection which provides a greater level of security and privacy.

\section{Proposed Methodology}
This section describes our proposed method to derive the protected fingerprint template. The overall design flow for the proposed method is displayed in Fig. 1. The proposed method consists of three major tasks which are displayed in rectangles in the block diagram. First, the input fingerprint image is preprocessed to extract the minutiae points by utilizing the thinned fingerprint image. Next, we form a nearest neighbor structure around each minutiae point using the ridge-based co-ordinate system and compute the ridge features from the thinned image and minutiae information. Thereafter, we apply Cantor pairing function to encode the ridge features uniquely. Finally, the random projection is applied onto the paired output to derive the protected template. In the verification phase, the same mechanism is followed to generate the protected query template from the query fingerprint and comparison is performed between the protected enrolled and protected query templates in the transformed (cancelable) domain.
	
	\begin{figure*}[t]
		\centering
		\includegraphics[width=\textwidth]{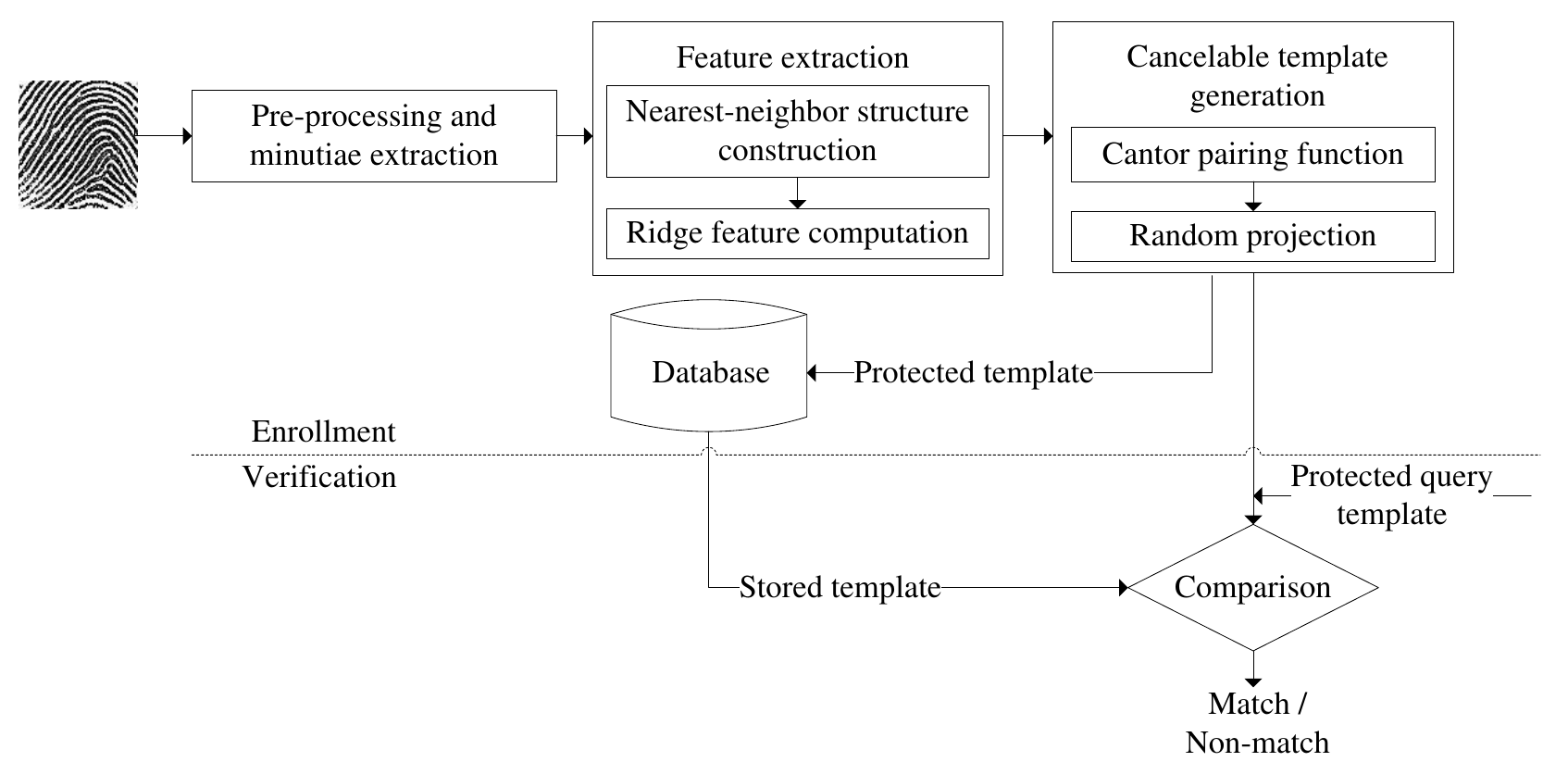}
		\vspace{-0.45cm}
		\caption{Block diagram of the proposed method}
		\vspace{-0.35cm}
		\label{fig1}\end{figure*}
	
	\subsection{Pre-processing and minutiae extraction}
	Fingerprint images may have different levels of contrast throughout the image. Pre-processing is performed to enhance the quality of input fingerprint image subsequently reducing the noise. In literature, different methods have been proposed to reduce noise and detect minutiae points from input fingerprint image. In this work, the pre-processing and extraction of minutiae points are performed by following the method presented in \cite{joshua}. The extracted minutiae points are denoted as follows:
	
	\begin{equation}
	\begin{split}
	V_{up}=\left \{ m_{i} \right \}_{i=1}^{n}         \\  
	m_{i}=\left ( x_{i}, y_{i}, \theta_{i} \right ) 
	\end{split}
	\end{equation}
	where, $V_{up}$ represents the set of untransformed (raw) minutia points detected from the input fingerprint and \textit{n} is the total number of minutiae points in $V_{up}$. The $i^{th}$ minutiae point is denoted by $m_{i}$ where $(x_{i},y_{i})$ and $\theta_{i}$ are the coordinate positions and orientation, respectively. Also, a thinned fingerprint image is obtained during preprocessing step which is further used for the invariant features extraction.
	
	\subsection{Nearest neighbor structure construction}
	We use minutiae information to create the nearest neighbor structure on the thinned fingerprint image. First, one of the minutiae point from \textit{$V_{up}$} is selected as a reference minutia. Next, the nearest neighbor structure is formed in the vicinity of reference minutiae point considering the ridge-based co-ordinate system as depicted in Fig. 2(a). In the ridge-based co-ordinate system, reference axis coincides with the orientation of the selected reference minutiae. Further, we divide the fingerprint region into `\textit{s}' sectors of equal angular width around the reference minutia in an anti-clockwise direction. In each sector, the nearest neighbor minutiae point is identified by selecting the minimum distance from the reference minutiae point. This procedure is followed for all minutiae points in $V_{up}$. It may be noted that if there is no minutia located in any of the sectors, we assign the nearest neighbor to be 0 in that sector. Further, we do not take into account the sectors with no minutiae point at the time of comparison. We consider eight sectors ($s$ = 8) in our method as shown in Fig. 2(a). In Fig. 2(a), $m_{2}$ is the nearest neighbor of reference minutiae $m_{1}$ in sector 3.
	
	\begin{figure*}[!htb]
		\begin{center}
			\subfigure[Nearest neighbor structure onto ridge-based coordinate system]{
				\includegraphics[width=0.42\textwidth,height=6cm,keepaspectratio]{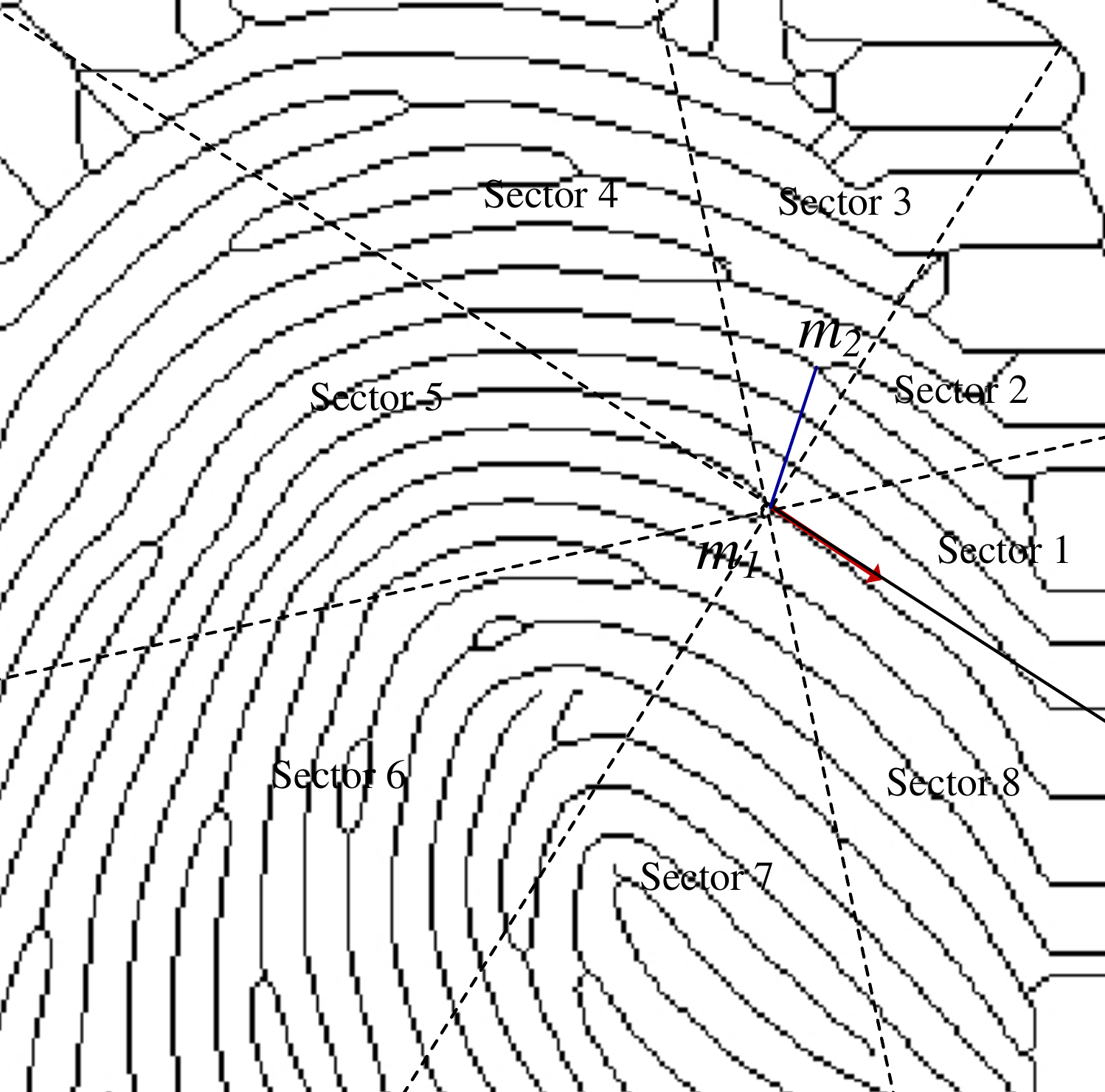}
			}
			\subfigure[Ridge-based feature computation]{
				\includegraphics[width=0.44\textwidth]{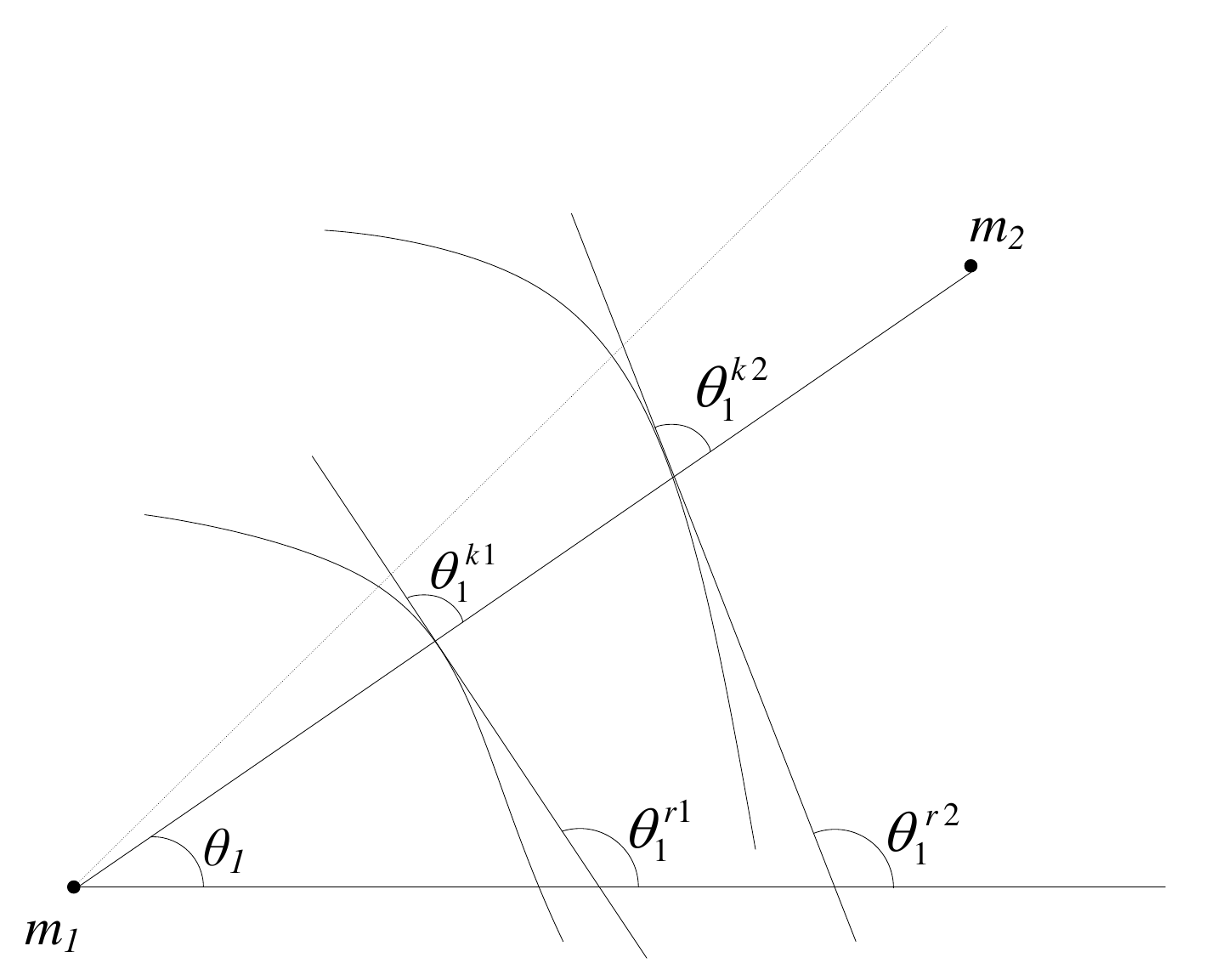}
			}
			\caption{Ridge feature extraction}
		\end{center}
	\end{figure*}
	
	\subsection{Ridge feature computation}
	The accuracy of the fingerprint-based verification system could be affected by translation, rotation and scale deformations produced while acquisition. Hence, it is necessary to compute invariant features from the input fingerprint image. In this work, we consider ridge count and average ridge orientation between the nearest neighbor minutiae and the reference minutiae in each sector as invariant features. To compute these features, the reference minutiae and the nearest neighbor minutiae points are identified first. Then, we compute the number of ridges along the straight line between these two minutiae in the thinned image and denote the ridge count in $j^{th}$ sector as $rc_{j}$. Figure 2(a) shows a descriptive example where ridge count between nearest neighbor minutiae ($m_{2}$) and the reference minutiae ($m_{1}$) is 2. To compute ridge orientation, a tangent is drawn at the intersection point of the line and ridge. Next, we measure the angle subtended by the tangent and straight lines between two minutiae points for each ridge crossing the straight line. For example, the orientation ($\theta_{1}^{k_{1}}$) of the first ridge in the first sector as shown in Fig. 2(b) is evaluated as:
	\begin{align*}
	\theta_{1}^{k_{1}}= \theta_{1}^{r_{1}} - \theta_{1}
	\end{align*}
	where $\theta_{1}$, denotes the slope of the line connecting nearest neighbor minutiae to the reference minutiae in the first sector. $\theta_{1}^{r_{1}}$ is the angle subtended by the tangent line from the first ridge crossing and the reference axis. In a similar manner, we calculate the orientation $\theta_{1}^{k_{2}}$ of the second ridge in the first sector and compute the mean ridge orientation for the first sector. The mean ridge orientation for the $j^{th}$ sector, denoted as $ro_{j}$ is calculated using Eq. (2).  
	
	\begin{equation}
	\resizebox{.9\hsize}{!}{
	$ro_{j}
	=\text{round} \left [\frac{\left ( \theta_{j}^{r_{1}} - \theta_{j} \right )+ \left ( \theta_{j}^{r_{2}} - \theta_{j} \right )  + ........+ 
		\left ( \theta_{j}^{r_{NR_{j}}} - \theta_{j} \right )}{NR_{j}} \right ]$}
	\end{equation}
	where, $NR_{j}$ represents the total number of ridges between the reference and nearest minutiae in the $j^{th}$ sector. Similarly, we find ridge count and mean ridge orientation for all minutiae and store it as $\left \langle \left \langle rc_{ij}, ro_{ij} \right \rangle_{j=1}^{s} \right \rangle_{i=1}^{n}$, where $s$ is the number of sectors and $n$ is the total number of minutiae.  
	
	\subsection{Cantor pairing function}
	The Cantor pairing function \cite{wiki,dover} is utilized to uniquely encode two natural numbers into a single natural number. Let $N$ = 0, 1, 2, 3,....  be the set of positive integers and $N \times N$ be the set of all ordered pairs of non-negative integers, a bijection from $N \times N$ to $N$ is called the Cantor pairing function which is defined as in Eq. (3).
	
	\noindent Consider a function:\quad  $\displaystyle \pi\colon \mathbb{N} \times \mathbb{N} \rightarrow \mathbb{N}$\\
	such that:
	\begin{equation}
	\pi (k_{1},k_{2}) := \frac{1}{2}(k_{1}+k_{2})(k_{1}+k_{2}+1)+k_{2}  \forall \left ( k_{1},k_{2} \right )\in \mathbb{N}^{2}  \label{eq:3}
	\end{equation}
	
	\vspace{0.1cm}
	We apply this function to encode ridge features. For each minutiae point, we compute the paired output of ridge features ($rc$ and $ro$) for each sector and stored in a 2D matrix as defined in Eq. (4). 
	\begin{eqnarray}
	CP\left ( i,j \right )=&\frac{1}{2}\left ( rc_{ij}+ro_{ij} \right )\left ( rc_{ij}+ro_{ij}+1 \right )+ ro_{ij}  \nonumber \\
	& \forall \ \ i\in [1,n] \ \text{and}\ j\in [1,s]
	\end{eqnarray}
	where, $rc_{i,j}$ and $ro_{i,j}$ represent the ridge features (i.e. ridge count and mean ridge orientation) between the nearest minutiae in the $j^{th}$ sector corresponding to $i^{th}$ reference minutiae. $CP\left ( i,j \right )$ is paired output of ridge features of the $j^{th}$ sector with respect to the $i^{th}$ reference minutiae. Next, we apply pointwise logarithm operation onto the paired outputs to obtain uniform distribution. Log function is defined in Eq. (5) and the base ($b$) of log function is chosen empirically (for details see Section III($C$)). For instance, if an input fingerprint image comprising $n$ minutiae points is divided into `$s$' sectors, the matrix $CP$ would result into $n\times s$ entries. After applying the log function, we obtain the log template (\textit{LT}) of size $n\times s$.
	\begin{equation}
	LT\left ( i,j \right ) = \log_{b}\left ( CP\left ( i,j \right ) \right )  \label{Eq:5}
	\end{equation}
	\subsection{Random projection}
	In order to derive non-invertible and revocable cancelable template, we perform random projection onto log template ($LT$). A random projection matrix ($RP$) of size $s\times t$ is derived using a random seed, $\kappa$ where $t < s$. Moreover, each of the entries of $RP$ is computed from a Gaussian independent and identical distribution (i.i.d.) with unit variance and mean is equal to zero. Now, each row of log template ($LT$) is projected onto random projection matrix ($RP$) to derive the cancelable template ($CT$) of size $n\times t$ as shown in Eq. (6). 
	\begin{equation}
	CT= LT \ldotp RP
	\end{equation}
	where, rank($RP$)=$r$. 
	
	In linear algebra, a theorem of linear systems \cite{aesbook} claims a unique solution when ranks of the coefficient matrix as well as the augmented matrix are same. Further, if rank becomes lower than the unknowns present, the linear system leads to infinite solutions as confirmed in the following proposition. Hence, $RP$ is one of those infinite solutions of Eq. (6).
	
	\vspace{0.2cm}
	\textbf{Proposition 1:} A linear non-homogeneous system of equations such as $CT= LT \ldotp RP$ including $s$ unknowns and n equations contain an infinite number of solutions.
	
	\vspace{0.1cm}
	\textbf{Proof:} Initially, we claim that $CT= LT \ldotp RP$ is solvable. It is evident from Eq. (6) that $CT$ is a linear combination of columns of $RP$, which states that $CT$ lies in the column space of $RP$. Hence, rank($RP$)=rank([$RP$ \ \ $CT$]). Due to same rank of coefficient matrix and augmented matrix, a solution exists for $CT= LT \ldotp RP$.
	
	Next, since rank($RP$)= $r < t$, there are infinitely many possible solutions to Eq. (6). The proposition illustrates that $CT$ is concealed among infinitely many possible solutions which become infeasible to an attacker even if he/she unveils $CT$ and $RP$. The attacker would not be able to achieve true biometric template as evaluation of pseudo-inverse results obsolete as shown in the following example:
	
	\noindent Example: Suppose, we have one row of log template and random projection matrix as follows:

\begin{equation}\resizebox{0.5\textwidth}{!}{$	
LT=\begin{bmatrix}
	2.5 & 1.3 & 3 & 4.51
\end{bmatrix} \text{and} \ \ RP=\begin{bmatrix}
	1.52 & -2.72 & 4.28 & -3.2 \\ 
	3 & -1.3 & 0.69 & -2.1 \\ 
	-0.76 & 1.36 & -2.14 & 1.6
\end{bmatrix}$}\nonumber
\end{equation}
	
Hence, rank($RP$)=2 and \begin{equation}
CT=LT \times RP = \begin{bmatrix}
-1.328 & -1.591 & -0.664
\end{bmatrix}\nonumber
\end{equation} is corresponding protected template. 
	
\noindent Let $_{i}^{p}\textrm{LT}$ represents the pseudo-inverse $LT$. $_{i}^{p}\textrm{LT}$ is computed as: 
	
\noindent $_{i}^{p}\textrm{LT} =RP^{\dagger}\cdot CT$=[-0.3579 \ 0.0993 \ 0.0240 \ 0.1927], \newline where, $RP^{\dagger}$ denotes pseudo-inverse of $RP$. Concurrently, we evaluate another solution manually, \newline ${_{i}^{p}\textrm{LT}}^{\ddagger}$=[0 \ 0 \ 0.3396 \ 0.8692]. Hence, it is verified that $_{i}^{p}\textrm{LT}\times RP=CT$  and ${_{i}^{p}\textrm{LT}}^{\ddagger} \times RP=CT$. 
	
This random projection based transformation guarantees the privacy and security of the proposed method. An imposter has no clue about $LT$ even if the protected template gets compromised. Further, if we consider the worst case of stolen $CT$ and $RP$, it would be very hard to retrieve $LT$ from infinitely many possible solutions. We illustrate this with a mathematical proof \cite{wsbook}:
	
	\vspace{0.2cm}
	\textbf{Proposition 2:} An underdetermined system of linear equations either contains an infinite number of solutions or become inconsistent. 
	
	\vspace{0.1cm}
	\textbf{Proof:} Consider this linear and underdetermined system, $CT= LT \ldotp RP$ (see Eq. (6)).
	
	We assume that $CT= LT \times RP$ has infinitely many solutions. Let $P$ be the $n \times s$ matrix,
	\[
	P=\begin{pmatrix}
	1 & 0 & \cdot & \cdot & \cdot & 0 & 0  \\ 
	0 & 1 & \cdot & \cdot & \cdot & 0 & 0  \\ 
	\cdot & \cdot & \cdot & \cdot & \cdot & \cdot & \cdot  \\ 
	\cdot & \cdot & \cdot & \cdot & \cdot & \cdot & \cdot  \\ 
	\cdot & \cdot & \cdot & \cdot & \cdot & \cdot & \cdot  \\  
	0 & 0 & \cdot & \cdot & \cdot & 1 & 0  \\ 
	\end{pmatrix}_{n \times s}\]
	
	We define $\hat{RP} :=P \ldotp RP$. Further we evaluate, 
	\[ \bar{LT}= \min_x \left \| P \ldotp RP-CT \right \|_{2}^{2} \]
	
	subject to the constraint $\left \| LT \ldotp RP - CT \right \|_{2}^{2} =0$. For evaluation of $\bar{LT}$, the \
	$(\lambda LT^{T} LT+P^{T} P)^{-1}$ must exist for all $\lambda>0$, this is non-trivial. Hence, $\bar{LT}$ achieved may look similar as $LT$, but it would not be identical.  
	
	The user's original information cannot be compromised even if an adversary obtains the stored fingerprint template because of the randomness present in the $RP$. The steps for cancelable template generation are summarized in Algorithm 1.

	\begin{algorithm}
		\caption{Cancelable template generation}
		\begin{algorithmic}[1]
			\renewcommand{\algorithmicrequire}{\textbf{Input:}}
			\renewcommand{\algorithmicensure}{\textbf{Output:}}
			\REQUIRE Log template ($LT_{n\times s}$), random projection matrix ($RP_{s\times t}$) \ \ \
			$\char`//$ Select a seed, $\kappa$  to generate random projection matrix $RP_{s\times t}$ where $t<s$
			\ENSURE  Cancelable template ($CT_{n\times t}$)
			\FOR {$i = 1$ to $n$ }
			\FOR {$j = 1$ to $t$}
			\STATE Initialize $sum$=0; 
			\FOR {$k = 1$ to $s$}
			\STATE  $sum$ $\leftarrow$  $sum$ + $LT(i,k) \times RP(k,j)$
			\ENDFOR    
			\STATE  $CT(i,j)$ $\leftarrow$ $sum$  
			\ENDFOR
			\ENDFOR
			\RETURN $CT_{n\times t}$
		\end{algorithmic} 
	\end{algorithm}
	
	\subsection{Comparison}
	The comparison between enrolled and query templates is performed in the protected domain to maintain secrecy. We compute local and global similarity to evaluate overall comparison score. We use Dice coefficient to measure the local similarity between the enrolled and query templates as utilized in \cite{mlce}. Finally, the likelihood of the enrolled and query template being the two fingerprint of the same subject is measured to compute global similarity score.
	
	\subsubsection{Local similarity score}
	Let us consider, the enrolled and query protected templates are denoted by $CT_{n\times t}$ and $QT_{m\times t}$ respectively where $m$ and $n$ represents the number of minutiae in the query and enrolled templates, respectively. To evaluate the local similarity score, each row of the \textit{CT} is cross-matched with all rows in \textit{QT} by computing the inner product of $CT(i,:)$ and $QT(j,:)$ where $i\in {1,2,\cdots, n}$ and $j \in {1, 2,\cdots, m}$. We obtain a similarity matrix $sim(i,j)\in{\Bbb R}^{n\times m}$ after applying the Eq. (7).
	
	\begin{equation}
	sim\left ( i,j \right ) = \frac{2 CT(i,:)\cdot QT(j,:)}{\left \| CT(i,:) \right \|^{2}+\left \| QT(j,:) \right \|^{2}} 
	\end{equation} 
	where $i \in [1,n]$ and $j \in [1,m]$.
	
	The $i^{th}$ row of $CT$ and the $j^{th}$ row of $QT$ are considered to be verifiable if and only if $sim(i,j)$ = max ([$sim(i,1), sim(i,2),.....,sim(i,n)$]) and $sim(i,j)$ = max ([$sim(1,j), sim(2,j),.....,sim(m,j)$]) are valid simultaneously. Therefore, each of the entries in similarity matrix is re-evaluated to eliminate double comparison in the following manner:
	
	Let,
	
	\begin{subequations}
		\begin{align}
		\Gamma_{CT} &=[\Gamma_{CT(1)},\ldots,\Gamma_{CT(i)}, \ldots \Gamma_{CT(n)}] \\
		& \text{where}, \ \Gamma_{CT(i)} = max (sim(i,1), \ldots, sim(i,m) \nonumber \\
		\intertext{and}
		\Gamma_{QT}&=[\Gamma_{QT(1)},\ldots,\Gamma_{QT(i)},\ldots,\Gamma_{QT(m)}] \\
		& \text{where}, \ \Gamma_{QT(j)} = max (sim(1,j), \ldots, sim(n,j) \nonumber
		\end{align}
	\end{subequations}
	be the maximum scores acquired for all minutiae in $CT$ and $QT$, respectively. Next, we construct a binary mask $A\in \left \{ 0,1 \right \}^{n \times m}$, which records the positions of the coinciding maxima;
	
	\begin{equation}
	A(i,j)= \delta(\Gamma_{CT(i)} ==  \Gamma_{QT(j)})
	\end{equation}
	where $\delta(\cdot)$ returns 1 when the nested condition is true and 0, otherwise. Hence, the filtered similarity matrix is represented by:
	\begin{equation}
	\hat{S}= sim \odot A
	\end{equation}
	where, $\odot$ represents element-wise multiplication.
	
	\subsubsection{Global similarity score}
	To perform overall comparison score between $CT$ and $QT$, the likelihood of $CT$ and $QT$ being two instances of the same fingerprint is measured. From the similarity matrix ($\hat{S}$) obtained in Eq. (11), we calculate the comparison score ($\mathfrak S$) with the following equation:
	\begin{equation}
	\mathfrak S =\frac{\sum_{i=1}^{n} \sum_{j=1}^{m} \widehat{S(i,j)}}{min(m,n)}
	\end{equation}
	Algorithm 2 describes the overall comparison procedure.
	
	\begin{algorithm}[!htbp]
		\caption{Comparison}
		\begin{algorithmic}[1]
			\renewcommand{\algorithmicrequire}{\textbf{Input:}}
			\renewcommand{\algorithmicensure}{\textbf{Output:}}
			\REQUIRE Cancelable template ($CT_{n\times t}$), Query template ($QT_{m\times t}$)
			\ENSURE  Comparison score ($match\_score$)
			\\ Initialize : $sim(n,m) \leftarrow 0 $, $S(n,m) \leftarrow 0 $
			
			\FOR {$i = 1$ to $n$}
			\STATE $RM1 \leftarrow CT[i]$
			\FOR {$j = 1$ to $m$}
			\STATE $RM2 \leftarrow QT[j]$
			\STATE $\char`//$ Evaluate similarity score as described in Eq. (7)
			\STATE $sim[i,j]$ = InnerProduct($RM1, RM2$)  
			\ENDFOR
			\ENDFOR
			\STATE $\Gamma_{CT} \leftarrow [\Gamma_{CT(1)},\ldots,\Gamma_{CT(i)}, \ldots \Gamma_{CT(n)}]$ \\ \text{where}, \ $\Gamma_{CT(i)} = max (sim(i,1), \ldots, sim(i,m)$
			\STATE $\Gamma_{QT} \leftarrow [\Gamma_{QT(1)},\ldots,\Gamma_{QT(i)},\ldots,\Gamma_{QT(m)}]$ \\ \text{where}, \ $\Gamma_{QT(j)} = max (sim(1,j), \ldots, sim(n,j)$  
			\FOR {$i = 1$ to $n$}
			\FOR {$j = 1$ to $m$}
			\STATE $S[i,j] \leftarrow (\Gamma_{CT}[i]=\Gamma_{QT}[j])$  $\char`//$\ Re-evaluate similarity score to avoid double comparison as described in Eq. (9)
			\ENDFOR
			\ENDFOR
			\STATE $sim \leftarrow sim\cdot^{\ast}S$  
			\STATE $match\_score(\mathfrak S)=\frac{sum(sim(:))}{min(m,n)}$  $\char`//$\ Evaluate Eq. (11)
		\end{algorithmic} 
	\end{algorithm}
		
	\section{Experimental results and analysis}
	In this section, we present the details of experimental design and results to illustrate the performance of the proposed method. We also analyze the effect of the different parameters as well as comparison with the existing approaches. 
	
	\subsection{Database selection}
	We have conducted our experiment on publicly available fingerprint databases FVC2002, FVC2004, and FVC2006 and each database contain four sets namely, DB1, DB2, DB3, and DB4 since most of the authors of biometrics research community utilize FVC databases \cite{fvc}. Each set of the first two databases comprise of 100 subjects with 8 images per subject. Each set of the FVC2006 database includes 140 subjects with 12 images per subject. 
	
	\subsection{Experimental design}
	In accordance with the ISO standard \cite{isoiec}, we use the following four metrics to evaluate the performance of our method:
	\begin{enumerate}
		\item FMR: The probability of getting a positive comparison decision for an imposter 
		\item FNMR: The probability of getting a negative comparison decision for a genuine user 
		\item GMR: Can be calculated as 1-FNMR
		\item EER: The error rate where FMR and FNMR hold equality
	\end{enumerate}
	The computation of these performance metrics involves the evaluation of genuine and imposter scores. Genuine score refers to the comparison of a fingerprint impression of a subject with the other impressions of the same subject, whereas imposter score is derived by comparing a fingerprint impression of each subject against the fingerprint impressions of all other subjects. Also, we have used standard FVC protocol and 1VS1 protocol to compute the performance of our method.
	
	In the 1VS1 protocol, the first fingerprint image of each subject is compared with the second fingerprint image of the same subject to compute the FNMR. To measure the FMR, the first image of each subject is compared with the first image of other subjects. This results to measure 100 genuine and $\comb{100}{2}$ =4950 imposter scores for each set of the FVC2002 and FVC2004 databases. For each set of FVC2006 database, 140 genuine and $\comb{140}{2}$=9730 imposter scores are computed. 
	
	In the FVC protocol, each fingerprint image of a subject is compared with the remaining fingerprint images of the same subject to compute the FNMR and to evaluate the FMR, the first fingerprint image of each subject is compared with the first fingerprint image of the different subjects. This results in providing $\comb{8}{2}\times100$=2800 genuine and $\comb{100}{2}$=4950 imposter scores computation for each set of FVC2002 and FVC2004 databases. For each set of FVC2006, $\comb{12}{2}\times140$=9240 genuine and $\comb{140}{2}$=9730 imposter scores are computed.
	
	\subsection{Validation of parameters}
	
	The proposed method utilizes two parameters to derive the protected fingerprint template. These parameters are: number of sectors ($s$) in the nearest neighbor structure [see section III(B)] and log-base value (\textit{b}) [see section III(D)]. In this section, we highlight the impact of these parameters on the performance of our approach. We have validated these parameters with respect to dataset DB1 of FVC2002, DB3 of FVC2004 and DB1 of FVC2006 using FVC protocol since they have good quality images.
	\begin{enumerate}
		\item \textit{Number of sectors ($s$)}: 
		After conducting pre-processing steps, we divide input fingerprint image into $s$ number of sectors with an equal angular width in the ridge-based co-ordinate system. To validate the parameter $s$, we have performed exhaustive testing considering the different angular widths with $15\degree$ interval. We have considered $s$ = 24, 12,\ldots and 4 corresponding to angular width $15\degree$, $30\degree$, \ldots and $90\degree$, respectively. To carry out this experiment, we have considered log-base value ($b$) as 1.2. The performance is measured with respect to EER and results are reported in Table 1. It has been observed that the method gives the best result for $s$ =8 on dataset DB1 of FVC2002, DB3 of FVC2004 and DB1 of FVC2006. The small value of $s$ degrades the performance of the method as the transformation becomes sensitive to noise. We also observe that EER increases for large values of $s$ as there are more sectors with 0 minutiae points. Hence, we have considered $s$ = 8 for all other experimental evaluations.
		
		\begin{table}
			\centering
			\caption{Number of sectors in nearest neighbor structure}
			\label{my-label}
			\begin{tabular}{|c|c|c|c|c|}
				\hline
				\multirow{2}{*}{\begin{tabular}[c]{@{}c@{}}Angular\\ width\end{tabular}} & \multirow{2}{*}{\begin{tabular}[c]{@{}c@{}}Number of \\ sectors ($s$)\end{tabular}} & \multicolumn{3}{c|}{EER (in \%)} \\ \cline{3-5} & 
				&  \begin{tabular}[c]{@{}c@{}}FVC2002\\ DB1\end{tabular} & \begin{tabular}[c]{@{}c@{}}FVC2004\\ DB3\end{tabular} & \begin{tabular}[c]{@{}c@{}}FVC2006\\ DB1\end{tabular} \\ \hline
				15 & 24 & 5.03  & 7.89 & 11.32 \\ \hline
				30 & 12 & 3.91  & 6.75 & 8.03  \\ \hline
				45 & 8  & 1.75  & 3.97 & 5.14  \\ \hline
				60 & 6  & 2.17  & 4.64 & 6.38  \\ \hline
				90 & 4  & 3.81  & 5.44 & 7.19  \\ \hline
			\end{tabular}
		\end{table}
		\item \textit{Log-base value ($b$)}: 
		We apply the log function onto the paired output derived using Cantor pairing function to obtain the uniform features distribution. We have conducted a number of experiments by considering the different values of $b$ = 1.1, 1.2, 1.3,..., 2 and measured the performance in terms of EER for dataset DB1 of FVC2002, DB3 of FVC2004 and DB1 of FVC2006 as reported in Table 2. The experimental evaluation illustrates that the method performs the best on $b$ = 1.2. We observe that small value of $b$ amplifies the distribution of paired output reducing EER. Also, EER gets increases as the discrimination between features of the different subjects gets reduced for high values of $b$. Therefore, we consider $b$=1.2 to evaluate the performance of our method.
	\end{enumerate}
	\vspace{-0.1cm}
	\begin{table}[!htb]
		\centering
		\caption{EER for different values of $b$}
		\label{my-label}
		\begin{tabular}{|c|c|c|c|}
			\hline
			\multirow{1}{*}{\begin{tabular}[c]{c@{}}Log-base\\ $(b)$\end{tabular}} &   \multicolumn{3}{c|}{EER (in \%)} \\ \cline{2-4} 
			& \begin{tabular}[c]{@{}c@{}}FVC2002\\ DB1\end{tabular} & \begin{tabular}[c]{@{}c@{}}FVC2004\\ DB3\end{tabular} & \begin{tabular}[c]{@{}c@{}}FVC2006\\ DB1\end{tabular} \\ \hline
			1.1 & 2.13   & 4.03  & 6.84   \\ \hline
			1.2 & 1.75   & 3.97  & 5.14   \\ \hline
			1.3 & 2.43   & 4.87  & 7.04   \\ \hline
			1.4 & 4.07   & 5.13  & 8.93   \\ \hline
			1.5 & 5.51   & 6.83  & 9.94   \\ \hline
			1.6 & 6.91   & 8.03  & 11.53  \\ \hline
			1.7 & 8.12   & 10.23 & 13.18  \\ \hline
			1.8 & 10.03  & 11.89 & 14.91  \\ \hline
			1.9 & 11.36  & 13.97 & 16.12  \\ \hline
			2   & 12.89  & 14.64 & 17.9   \\ \hline
		\end{tabular}
	\end{table}
	
	\subsection{Performance evaluation}
	To measure the performance of our method, we have conducted two sets of experiments. We evaluate the performance under the same key and different key scenario in the first and second set of experiments, respectively. Each experiment is conducted 10 times, and the average performance of 10 trials is reported in the paper.
	
	\subsubsection{Same key scenario}
	This scenario represents the situation in practice when an imposter unveils the random projection matrix ($RP$). We have evaluated this scenario by assigning the same $RP$ to each user present in the database. The ROC curves for FVC2002, FVC2004, and FVC2006 databases with FVC and 1VS1 protocols are shown in Fig. 3, Fig. 4 and Fig. 5, respectively.  
	
	\textit{FVC2002}: For FVC2002 database, we achieve an EER of 1.75, 0.98, 4.02, and 3.74 for DB1, DB2, DB3, and DB4 datasets, respectively using FVC protocol. For 1VS1 protocol, we obtain an EER of 0, 0.13, 3.39, and 3.02 for DB1, DB2, DB3, and DB4, respectively. Out of all FVC2002 datasets, the method exhibit low EER on DB1 and DB2 for both protocols due to the presence of more number of good quality images as compared to other datasets of FVC2002. Further, \nth{1} and \nth{2} images of a subject in FVC2002 DB2 are acquired in the same session and have less variation and distortion than the other six images. However, images in DB3 and DB4 dataset of FVC2002 contain relatively poor quality images with less number of minutiae points as compared to dataset DB1 and DB2. As a result, we achieve high EER for DB3 and DB4 datasets under both protocols. 
	
	\begin{figure*}[!h]
		\includegraphics[width=\textwidth, height=0.37\textheight]
		{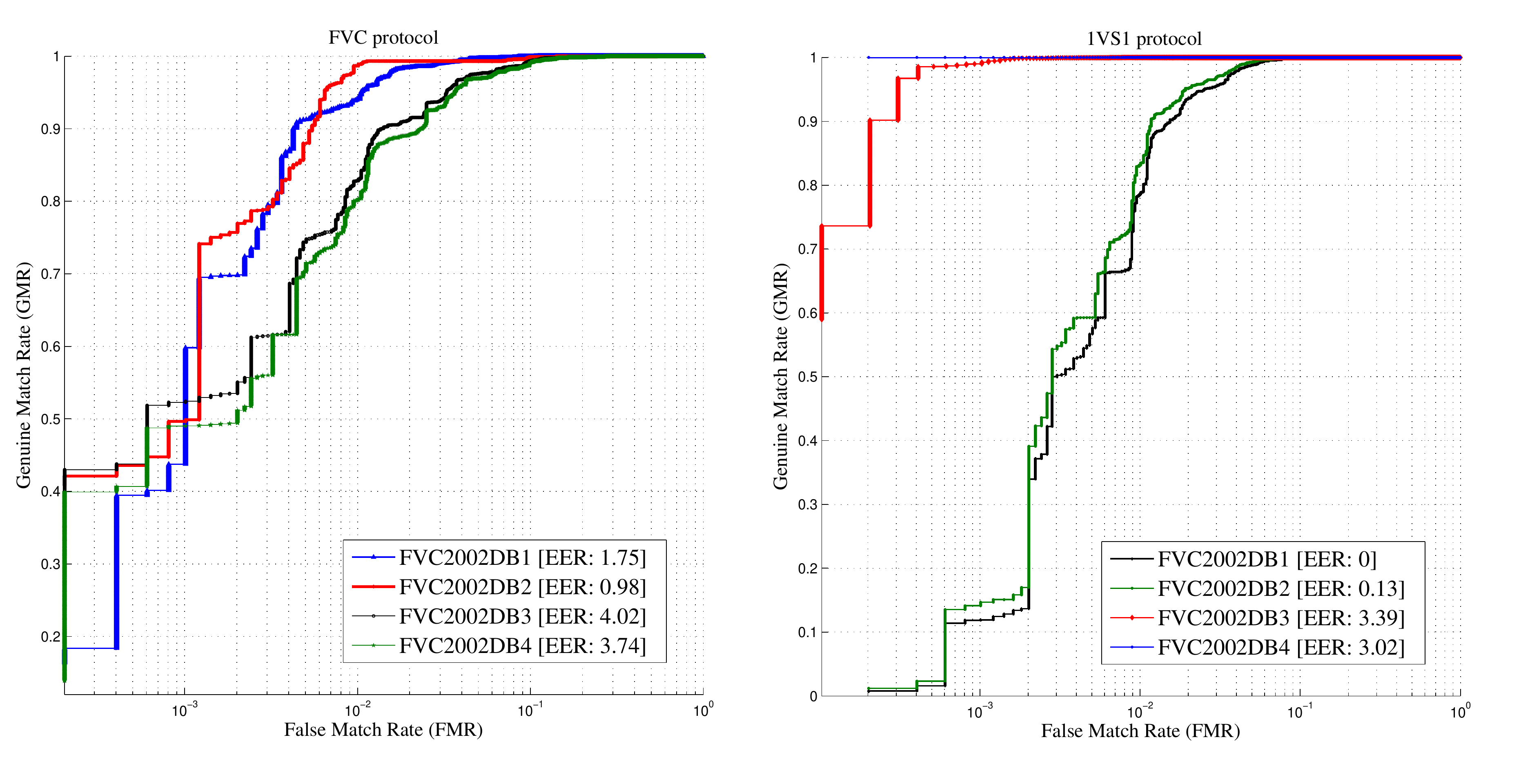}
		\caption{ROC curves for FVC2002 datasets under FVC and 1VS1 protocol}
	\end{figure*}
	
	\textit{FVC2004}: For FVC2004 database, we achieve an EER of 4.38, 6.59, 3.97, and 3.16 for DB1, DB2, DB3, and DB4 datasets, respectively using FVC protocol. In the 1VS1 protocol, we obtain an EER of 4.02, 5.77, 3.88, and 3.04 for DB1, DB2, DB3, and DB4, respectively. The method performs better on DB4, out of all FVC2004 datasets in both protocols. However, the method gives a high value of EER on DB2 for both protocols as the first two images of the DB2 dataset are heavily distorted. Moreover, the small overlap area corresponding to the images of stored and query templates is another reason for less accuracy on FVC2004 DB2. For example, if we consider the images of stored and query template, as 96\_1.tif and 96\_2.tif, the genuine verification attempt fails. This is because 96\_1.tif contains the region below the core point, whereas 96\_2.tif contains region above the core point. The absence of overlapping area causes very few or zero corresponding minutiae from the stored and query templates. Since the proposed system relies on minutiae neighborhood, the lack of corresponding minutiae pair from the stored and query template pair causes comparison trial to fail. We achieve high EER for all four datasets of the FVC2004 database since all the users were requested to put deliberate perturbations at the time of acquisition \cite{fvc}.
	\begin{figure*}[!h]
		\includegraphics[width=\textwidth, height=0.37\textheight]
		{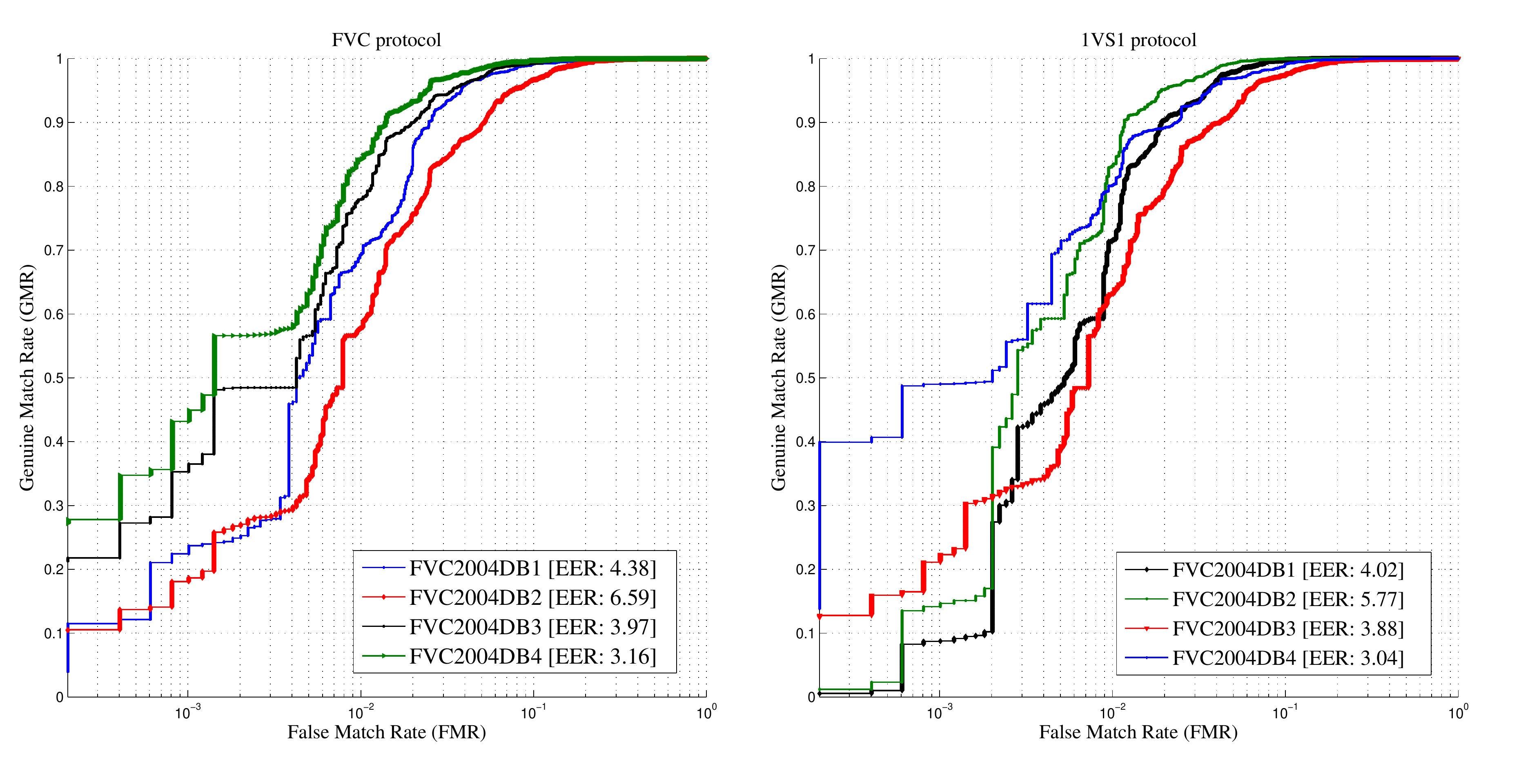}
		\vspace{-0.75cm}
		\caption{ROC curves for FVC2004 datasets under FVC and 1VS1 protocol}
	\end{figure*}
	
    \textit{FVC2006}: For FVC2006 dataset, we achieve an EER of 5.14, 0.14, 1.63, and 0.49 for DB1, DB2, DB3, and DB4, respectively using FVC protocol. For 1VS1 protocol, we obtain an EER of 3.8, 0.09, 2.02, and 1.03 for DB1, DB2, DB3, and DB4, respectively. All of the four datasets are selected among the heterogeneous populations (i.e., manual workers and elderly people) allowing most difficult fingerprints according to quality index with explicit distortions such as large amounts of rotation and displacement, wet/dry impressions, etc. The dataset DB1 contains small sized poor quality images with missing minutiae. Therefore, the method produces high EER on the DB1 dataset. The method performs optimally on the DB2 dataset for both protocols due to the presence of relatively good quality images. Dataset DB3 and DB4 consist of more number of poor quality images in comparison to dataset DB2. Therefore, it is observed that the performance of the method degrades for DB3 and DB4 of the FVC2006 dataset.
	
	\begin{figure*}[!h]
		\includegraphics[width=\textwidth, height=0.37\textheight]
		{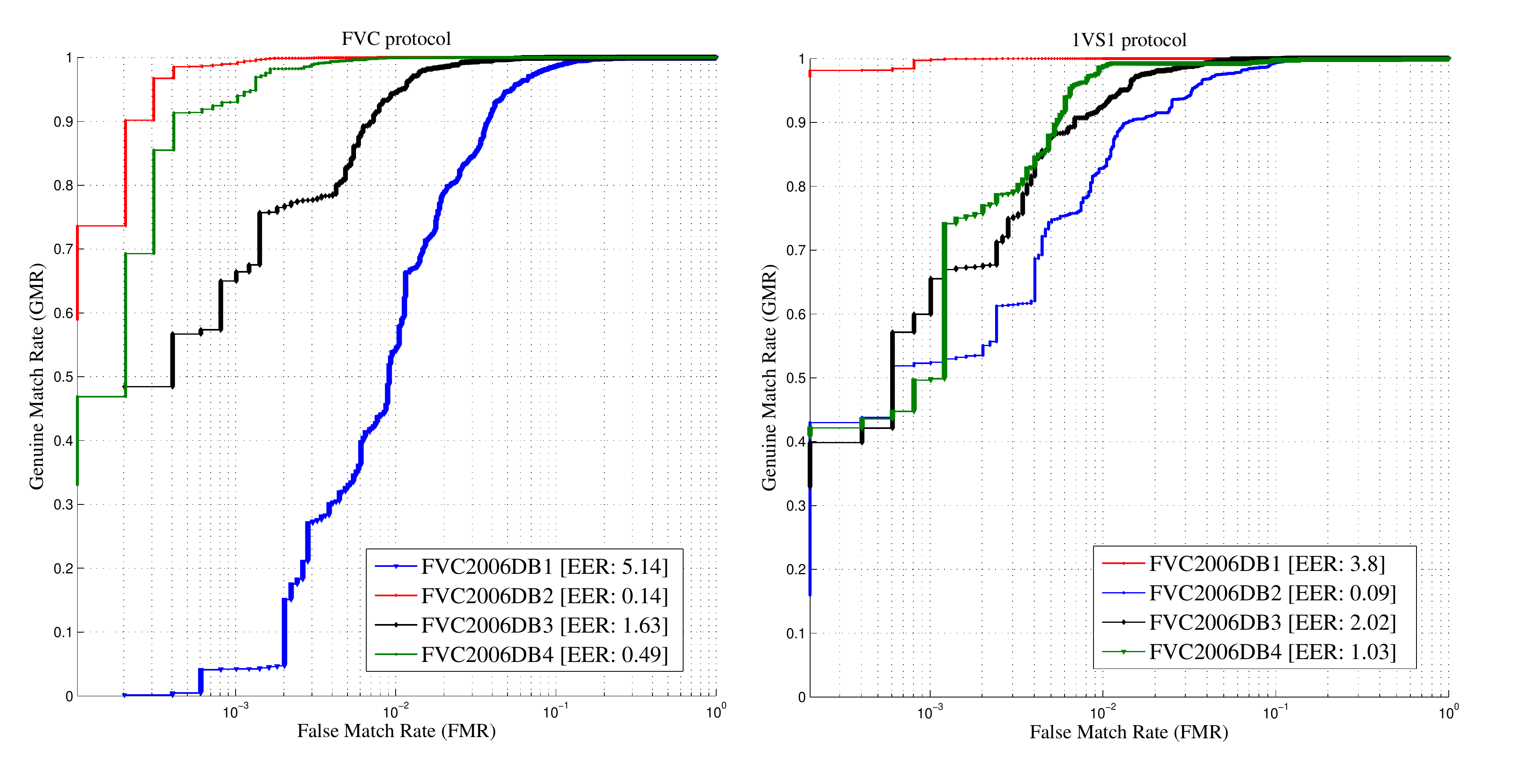}
		\vspace{-0.75cm}
		\caption{ROC curves for FVC2006 datasets under FVC and 1VS1 protocol}
	\end{figure*}
	
	Further, we also observe that the proposed method performs better with 1VS1 protocol compared to standard FVC protocol. The reason lies in the number of genuine verification attempts. In case of 1VS1 protocol, the first two images of the same user are utilized, whereas all eight images from each user are utilized in the genuine verification for the standard FVC protocol. However, we achieve high EER for 1VS1 protocol as compared to FVC protocol for the DB3 and DB4 datasets of the FVC2006 database since the first two images are noisy and involve non-overlapping regions. 
	
	\subsubsection{Different key scenario}
	In the second set of experiment, we assign the different projection matrices to the different users by altering the seed value and test our method for both the protocols. For FVC2002 dataset, our method performs ideal for all datasets (EER = 0) with both protocols. Moreover, we achieve an EER of 0 for DB1 and DB2 datasets of FVC2004. Datasets DB3 and DB4 consist of more number of poor quality images with very few or missing minutiae in comparison to datasets DB1 and DB2. For DB3 and DB4 dataset, the method gives EERs of 0.08 and 0.03, respectively. For FVC2006 dataset, we achieve an EER close to 0 for DB1, DB2, DB3, and DB4 using FVC and 1VS1 protocols. Therefore, it is evident that the performance of the method in the different key scenario is almost ideal for all datasets.
	
	\subsection{Comparison of with and without transformation}
	To perform a fair comparison, the verification performance of the proposed cancelable biometric system is analyzed with respect to the baseline biometric system (i.e., original ridge features). Further, a comparison process relying on the original minutiae should be taken into account, since the employed ridge-based representation is already part of the process generating the proposed protected templates. Therefore, we compare the performance of the method under three scenarios i.e. original minutiae comparison, original ridge features and protected templates.  In this experiment, first we compute the performance for original minutiae comparison based on adaptive image enhancement method proposed by Bartunek et al. \cite{tip}. The approach involves publicly available Bozorth3 minutiae matcher \cite{bozorth} from NIST to evaluate the performance. Next, we compute the performance using original ridge features of the query and stored templates. Further, we apply the proposed approach to derive cancelable template and compare the stored and query templates in the transformed domain. Table 3 reports the EERs obtained from the original minutiae comparison, original ridge features and the cancelable (protected) fingerprint template for the different datasets. It has been observed that the proposed ridge-based computation outperforms the original minutiae comparison since Bozorth3 does not perform well for poor quality fingerprint images with fewer minutiae points. Also, Bozorth3 is not robust against the alignment, and scale deformations present between the stored and query templates. Therefore, it is evident that the proposed method outperforms than the Bozorth3 matcher. 
	
	For FVC protocol, the reported results in Table 3 exhibit that there is a minor degradation of 0.19\%, 0.15\%, 0.05\% and 0.07\% in the performance for DB1, DB2, DB3, DB4 of FVC2002, 0.053\%, 0.07\%, 0.033\%, 0.04\% for DB1, DB2, DB3, DB4 of FVC2004 and 0.04\%, 0.14\%, 0.037\%, 0.32\% for DB1, DB2, DB3, DB4 of FVC2006, respectively with reference to original ridge features. For 1VS1 protocol, the reported results in Table 4 indicate that the performance is degraded by 0\%, 0.85\%, 0.08\% and 0.09\% for DB1, DB2, DB3, DB4 of FVC2002, 0.05\%, 0.11\%, 0.04\%, 0.05\% for DB1, DB2, DB3, DB4 of FVC2004 and 0.048\%, 2.0\%, 0.09\%, 0.17\% for DB1, DB2, DB3, DB4 of FVC2006, respectively with reference to original ridge features. Therefore, we can conclude that performance degradation produced by the transformation is very low.
	
	\begin{table*}[!htbp]
		\centering
		\caption{Baseline comparison for FVC protocol}
		\label{my-label}
		\begin{tabular}{|c|c|c|c|c|c|c|c|c|c|c|c|c|}
			\hline
			\multirow{3}{*}{Dataset} & \multicolumn{12}{c|}{EER} \\ \cline{2-13} 
			& \multicolumn{4}{c|}{\begin{tabular}[c]{@{}c@{}}Original minutiae\\ comparison\end{tabular}} & \multicolumn{4}{c|}{\begin{tabular}[c]{@{}c@{}}Without cancelable\\ transformation\end{tabular}} & \multicolumn{4}{c|}{\begin{tabular}[c]{@{}c@{}}With cancelable\\ transformation\end{tabular}} \\ \cline{2-13} 
			& DB1 & DB2 & DB3 & DB4 & DB1 & DB2 & DB3 & DB4 & DB1 & DB2 & DB3 & DB4 \\ \hline
			FVC2002 & 2.8 & 2.3 & 6.5 & 3.9 & 1.47 & 0.89 & 3.81 & 3.49 & 1.75 & 0.98 & 4.02 & 3.74 \\ \hline
			FVC2004 & 9.6 & 5.9 & 6.2 & 6.6 & 4.14 & 6.12 & 3.84 & 3.03 & 4.38 & 6.59 & 3.97 & 3.16 \\ \hline
			FVC2006 & 5.2 & 1.39 & 2.91 & 1.27 & 4.93 & 0.12 & 1.57 & 0.37 & 5.14 & 0.14 & 1.63 & 0.49 \\ \hline
		\end{tabular}%
	\end{table*}
	
	\begin{table*}[!htbp]
		\centering
		\caption{Baseline comparison for 1VS1 protocol}
		\label{my-label}
		\begin{tabular}{|c|c|c|c|c|c|c|c|c|c|c|c|c|}
			\hline
			\multirow{3}{*}{Dataset} & \multicolumn{12}{c|}{EER} \\ \cline{2-13} 
			& \multicolumn{4}{c|}{\begin{tabular}[c]{@{}c@{}}Original minutiae\\ comparison\end{tabular}} & \multicolumn{4}{c|}{\begin{tabular}[c]{@{}c@{}}Without cancelable\\ transformation\end{tabular}} & \multicolumn{4}{c|}{\begin{tabular}[c]{@{}c@{}}With cancelable\\ transformation\end{tabular}} \\ \cline{2-13} 
			& DB1 & DB2 & DB3 & DB4 & DB1 & DB2 & DB3 & DB4 & DB1 & DB2 & DB3 & DB4 \\ \hline
			FVC2002 & 0.91 & 1.02 & 4.3 & 3.89 & 0 & 0.07 & 3.13 & 2.77 & 0 & 0.13 & 3.39 & 3.02 \\ \hline
			FVC2004 & 4.65 & 6.30 & 4.72 & 3.95 & 3.81 & 5.19 & 3.69 & 2.89 & 4.02 & 5.77 & 3.88 & 3.04 \\ \hline
			FVC2006 & 4.87 & 1.04 & 2.65 & 2.83 & 3.62 & 0.03 & 1.83 & 0.88 & 3.8 & 0.09 & 2.02 & 1.03 \\ \hline
		\end{tabular}%
	\end{table*}
	
	\subsection{Comparison with the existing cancelable biometric approaches}
	\begin{sloppypar}
		The approaches in \cite{pairp,boult,lmi,ditom,byang,blind,hadamard} used FVC 2002 database to evaluate the performance of their method using standard FVC protocol. Further, Wong et al. \cite{mlc} also evaluated the performance on DB1 of FVC2004. In addition to each dataset of FVC2002, Ferrara et al. \cite{pmcc,2pmcc} evaluated their methods on DB1 of FVC2004 and DB2 of FVC2006. 
		The authors, Yang et al. \cite{byang} and Wang et al. \cite{blind} evaluated the performance on DB2 of FVC2002 with the 1VS1 protocol. Ferrara et al. \cite{pmcc,2pmcc} also evaluated their methods on DB2 of FVC2006 and each datasets of FVC2002 database for 1VS1 protocol. Therefore, we compare our method with these current state-of-the-art approaches \cite{pairp,boult,lmi,ditom,byang,mlc,pmcc,2pmcc,blind,hadamard} in the literature. Table 5 and 6 summarizes the comparison in terms of EER on different FVC datasets for 1VS1 and FVC protocols, respectively. From Table 5, we observe that the best result reported in existing literature is EER = 0, 0.02, 3.43, 3.37 and 0.03 for FVC2002DB1, FVC2002DB2, FVC2002DB3, FVC2002DB4 and FVC2006DB2, respectively, whereas our approach yields EER of 0, 0.13, 3.39, 3.02 and 0.09 for FVC2002DB1, FVC2002DB2, FVC2002DB3, FVC2002DB4 and FVC2006DB2, respectively. From Table 6, we observe that the best result reported in existing literature is EER = 1, 0.99, 5.24, 4.84, 10.36 and 0.17 for FVC2002DB1, FVC2002DB2, FVC2002DB3, FVC2002DB4, FVC2004DB1 and FVC2006DB2, respectively, whereas our approach gives EER of 1.75, 0.98, 4.02, 3.74, 4.38 and 0.14 for FVC2002DB1, FVC2002DB2, FVC2002DB3, FVC2002DB4, FVC2004DB1 and FVC2006DB2, respectively. However, we can observe that the performance of the proposed method for the DB2 dataset of FVC2002 and FVC2006 is slightly lower than \cite{pmcc} in 1VS1 protocol and the EER of FVC2002DB1 is lower than that of Wang et al. \cite{hadamard}, but it is comparable. Therefore, it is evident from the reported results that the performance of our proposed approach performs better over the existing template protection techniques.
	\end{sloppypar}
	
	\begin{table}[!htb]
		\centering
		\caption{Performance comparison with existing cancelable approaches for 1VS1 protocol (values in percentages)}
		\label{my-label}
		\begin{tabular}{|p{0.6cm}|p{0.7cm}|c|c|c|c|}
			\hline
			\multicolumn{2}{|c|}{\multirow{3}{*}{Datasets}} & \multicolumn{4}{c|}{Methods} \\ \cline{3-6} 
			\multicolumn{2}{|c|}{} & \multicolumn{1}{c|}{\multirow{2}{*}{\begin{tabular}[c]{@{}c@{}}Yang \\ et al. \cite{byang}\end{tabular}}}    & \multicolumn{1}{c|}{\multirow{2}{*}{\begin{tabular}[c]{@{}c@{}}Ferrara \\ et al. \cite{pmcc}\end{tabular}}}   & 
			\multicolumn{1}{c|}{\multirow{2}{*}{\begin{tabular}[c]{@{}c@{}}Wang \\ et al. \cite{blind}\end{tabular}}}  & \multicolumn{1}{c|}{\multirow{2}{*}{\begin{tabular}[c]{@{}c@{}}Proposed \\ method \end{tabular}}} \\
			\multicolumn{2}{|c|}{} & \multicolumn{1}{c|}{} & \multicolumn{1}{c|}{} & \multicolumn{1}{c|}{} & \multicolumn{1}{c|}{} \\ \hline
			\multirow{4}{*}{\begin{tabular}[c]{@{}c@{}}FVC \\ 2002\end{tabular}} 
			& DB1 & - & 0 & 3 & 0 \\ \cline{2-6} 
			& DB2 & 0.72 & 0.02 & 2 & 0.13 \\ \cline{2-6} 
			& DB3 & - & 3.43 & 7 &3.39 \\ \cline{2-6} 
			& DB4 & - & 3.37 & - & 3.02 \\ \hline
			\begin{tabular}[c]{@{}c@{}}FVC \\ 2006\end{tabular} 
			& DB2 & - & 0.03  & - & 0.09 \\ \hline
		\end{tabular}
	\end{table}
	
	\begin{table*}[t]
		\centering
		\caption{Performance comparison with existing cancelable approaches for FVC protocol (values in percentages)}
		\label{my-label}
		\resizebox{\textwidth}{!}{%
			\begin{tabular}{|c|c|c|c|c|c|c|c|c|c|c|c|c|}
				\hline
				\multicolumn{2}{|c|}{\multirow{3}{*}{Datasets}} & \multicolumn{11}{c|}{Methods} \\ \cline{3-13} 
				\multicolumn{2}{|c|}{} & \multicolumn{1}{c|}{\multirow{2}{*}{\begin{tabular}[c]{@{}c@{}}Ahmad \\ et al. \cite{pairp}\end{tabular}}} &  \multicolumn{1}{c|}{\multirow{2}{*}{\begin{tabular}[c]{@{}c@{}}Wang\\ et al. \cite{ditom}\end{tabular}}} & \multicolumn{1}{c|}{\multirow{2}{*}{\begin{tabular}[c]{@{}c@{}}Lee et al. \\ \cite{lmi}\end{tabular}}} & \multicolumn{1}{c|}{\multirow{2}{*}{\begin{tabular}[c]{@{}c@{}}Wong\\ et al. \cite{mlc}\end{tabular}}} & \multicolumn{1}{c|}{\multirow{2}{*}{\begin{tabular}[c]{@{}c@{}}Yang \\ et al. \cite{byang}\end{tabular}}} & \multicolumn{1}{c|}{\multirow{2}{*}{\begin{tabular}[c]{@{}c@{}}Boult \\ et al. \cite{boult}\end{tabular}}} & \multicolumn{1}{c|}{\multirow{2}{*}{\begin{tabular}[c]{@{}c@{}}Ferrara\\ et al. \cite{pmcc}\end{tabular}}} & \multicolumn{1}{c|}{\multirow{2}{*}{\begin{tabular}[c]{@{}c@{}}Ferrara \\ et al. \cite{2pmcc}\end{tabular}}} & \multicolumn{1}{c|}{\multirow{2}{*}{\begin{tabular}[c]{@{}c@{}}Wang \\ et al. \cite{blind}\end{tabular}}} & \multicolumn{1}{c|}{\multirow{2}{*}{\begin{tabular}[c]{@{}c@{}}Wang \\ et al. \cite{hadamard}\end{tabular}}} &	\multicolumn{1}{c|}{\multirow{2}{*}{\begin{tabular}[c]{@{}c@{}}Proposed \\ method\end{tabular}}} \\
				\multicolumn{2}{|c|}{} & \multicolumn{1}{c|}{} & \multicolumn{1}{c|}{} & \multicolumn{1}{c|}{} & \multicolumn{1}{c|}{} & \multicolumn{1}{c|}{} & \multicolumn{1}{c|}{} & \multicolumn{1}{c|}{} & \multicolumn{1}{c|}{} & \multicolumn{1}{c|}{} & \multicolumn{1}{c|}{} & \multicolumn{1}{c|}{}  \\ \hline
				\multirow{4}{*}{\begin{tabular}[c]{@{}c@{}}FVC\\ 2002\end{tabular}} 
				& DB1 & 9  & 3.5  & 3.4 & 4.69  & -    & 2.1 & 1.88 & 3.3 & 4 & 1 & 1.75 \\ \cline{2-13} 
				& DB2 & 6  & 4    & -   & 5.03  & 4.53 & 1.2 & 0.99 & 1.8 & 3 & 2 & 0.98 \\ \cline{2-13} 
				& DB3 & 27 & 7.5  & -   & -     & -    &  -  & 5.24 & 7.8 & 8.5 & 5.2 & 4.02 \\ \cline{2-13} 
				& DB4 & -  & -    & -   & -     & -    &  -  & 4.84 & 6.6 & - & - & 3.74 \\ \hline
				\begin{tabular}[c]{@{}c@{}}FVC\\ 2004\end{tabular} 
				& DB1 & - & - & - & 10.36 & -  & - & - & 6.3 & - & - & 4.38 \\ \hline
				\begin{tabular}[c]{@{}c@{}}FVC\\ 2006\end{tabular} 
				& DB2 & - & - & - & - & - & -  & 0.17 & 0.3 & - & - & 0.14 \\ \hline
			\end{tabular}%
		}
		\begin{tablenotes}\small
			\item $``-"$ indicates that the author(s) have not reported the results or results are reported for partial dataset, in their work.
		\end{tablenotes}
	\end{table*}
	
	\section{Security analysis}
	The security of the derived protected template is guaranteed when an adversary has no information about the transformation. If an adversary unveils any information about cancelable transformation, the security of the proposed system is guaranteed by three factors: non-invertibility, revocability, and diversity. In this section, we analyze our method with respect to these three contexts.
	
	\subsection{Non-invertibility}\label{nonin}
	The term, non-invertibility refers that it should be computationally infeasible to derive the original fingerprint template from the protected template. Note that a randomized projection matrix ($RP$) is utilized to generate a cancelable template from the log template ($LT$). To meet the non-invertibility requirement, we have adopted a reference architecture proposed by Breebaart et al. \cite{iso}. Figure 6 shows the reference architecture where the protected template ($CT$), random projection matrix ($RP$), and the parameters ($s$, $b$) can be presumed as pseudo-identity, auxiliary data and supplementary data, respectively.
	
	\begin{figure*}[!htbp]
		\centering
		\includegraphics[width=\textwidth]{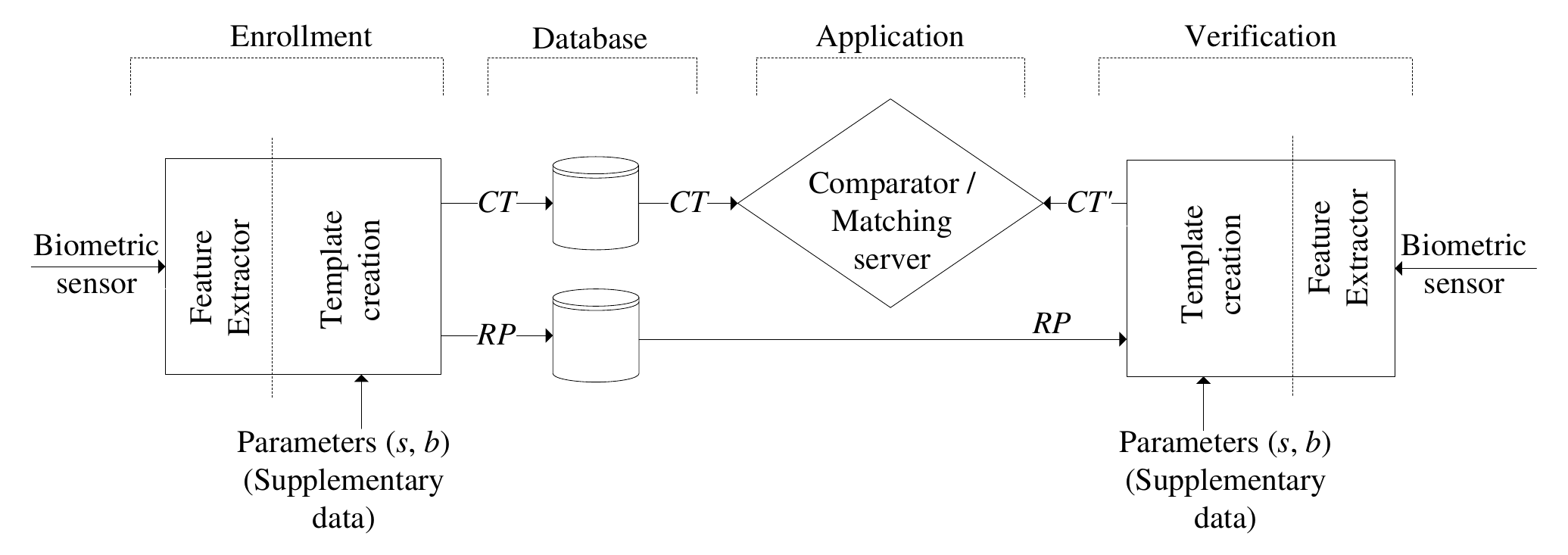}
		\caption{Reference architecture for the creation, storage and verification of the protection template}
	\end{figure*}
	
	In the reference architecture, the protected template ($CT$) is derived at the enrollment phase. The biometric sample, ridge features and the parameter ($s$, $b$) are destroyed after the successful verification of stored protected and query protected templates. Due to privacy preservation, it may have been either issued for a limited period or may require revocation when compromised. Moreover, the biometric characteristics may get affected due to aging effects. Hence, it requires renewal after a validity period regulated through watch list. The protected template $CT$ along with the $RP$ and supplementary data ($s,b$) are stored in the database. During verification, a protected query template ($CT^{'}$) is generated from the issued $RP$, biometric sample and the parameters ($s$, $b$). Next, the stored protected template ($CT$) and the query protected template ($CT^{'}$) are forwarded to a comparator/matching server via the communication interface to verify the identity. In this section, we analyze the criterion of non-invertibility with three different architectural components i.e. database, matching server and communication interface for information exchange. 
	
	\subsubsection{Compromised database}
	In this scenario, an attacker can reveal the database i.e. protected template ($CT$) and the random projection matrix ($RP$). On the possession of this information, the attacker would not be able to retrieve the log template ($LT$) since the size of $RP$ is $s\times t$ where $t<s$ and the entries of $RP_{s\times t}$ are independent and identically distributed (i.i.d.) Gaussian random variables. Evaluation of $LT_{n\times s}$ from $CT_{n\times t}$ results to find a solution for underdetermined system because it is hard to find $s$ unknowns from $t$ linearly independent equations where $t<s$ (see Proposition 1-2 in Section III-E). Further, it has also been proved in Du et al. \cite{ksecure} that if the projection matrix follows the condition $t\leq \frac{s}{2}$ and entries in $RP$ are i.i.d., it is very hard to find the $LT$ from $CT$. Moreover, even if the attacker achieves supplementary information ($s, b$), it would be infeasible to unveil the $LT$ as analyzed in the third scenario i.e. compromised communication interface.
	
	\subsubsection{Compromised matching server}
	Let us assume that an attacker unveils matching server i.e. stored protected template ($CT$) and query protected template ($CT^{'}$). Next, an attacker tries to evaluate $LT$ by correlating the information contained in $CT$ and $CT^{'}$. In this situation, an attacker would not be able to retrieve $LT$ since he does not have any information about the $RP$.
	
	\subsubsection{Compromised communication interface}
	In this scenario, an attacker may have control over communication interface between the database and matching server. In this situation, the adversary would be able to estimate the stored protected template ($CT$), query protected template ($CT^{'}$) and the random projection matrix ($RP$). On the possession of these information, the attacker may utilize $CT$ and $RP$, or $CT^{'}$ and $RP$ to retrieve the log template ($LT$). This situation is identical to the first scenario i.e. compromised database. Further, the attacker may correlate $CT$ and $CT^{'}$ to evaluate $LT$. This situation is same as the second scenario i.e. compromised matching server.
	
	Further, we assume that the attacker unveils the approximate $LT$ by applying known key distinguishing attack. In this situation, the imposter tries to estimate $CP$ or approximate $CP$ using the value of parameter $b$. However, it would not be possible to retrieve original ridge features since the inversion involves the computation of a square root which gives one to many correspondences as defined in Eq. 12(a)-(d).
	
	\begin{subequations}
		\begin{align}
		w&=\left \lfloor \frac{\sqrt{8\ CP +1}-1}{2} \right \rfloor \label{first} \\
		t&=\frac{w^{2}+w}{2}\label{second}\\
		ro&=CP-t \label{third}   \\
		rc&=w-ro \label{fourth}
		\end{align}
	\end{subequations}
	where, $rc$, $ro$ and $CP$ represent the ridge count, mean ridge orientation and transformed paired output, respectively. $w$ and $t$ are the intermediate values in the calculation and $\left \lfloor \hspace{0.25mm} \right \rfloor$ is the floor function. Hence, it would not be possible to invert $CP$ to attain the original ridge features. Therefore, it can be stated that our method preserves the criterion of non-invertibility.

	\subsection{Revocability}
	The term revocability refers to the design of a new protected template if stored template gets leaked. The newly generated template should be adequately dissimilar to the compromised one. In this work, a new protected template can be issued just by altering $RP$. To ensure the potent revocability, the biometric templates that are derived by applying different $RP$s for the same user in different applications, should not be able to verify each other. Here, the random projection is motivated by the Johnson-Lindenstrauss (JL) lemma described in \cite{jlproof}. The lemma states that: \\
	For any $0 < \epsilon <1$ and an integer $k$, let $t$ be a positive integer such that $t \geq t_{0}= \mathcal{O}$ ($\epsilon^{-2}log\ k$). For any set $B$ of $k$ points in $\Re^{s}$, there exists a map $f: \Re^{s}\rightarrow \Re^{t}$ such that: for all $u, v \in B$, 
	
	\begin{equation}
	(1-\epsilon) \left \| u-v \right \|^{2}\leqslant \left \| f(u)-f(v) \right \|^{2} \leq (1+\epsilon) \left \| u-v \right \|^{2}
	\end{equation}
	where, $u$ and $v$ are two randomly derived vectors in the $s$-dimensional Euclidean space, $u,v \in \Re^{s}$. For inner-product based similarity, it states that: 
	
	\begin{center}
		\begin{math}
		\frac{u.v}{\left \| u \right \|.\left \| v \right \|}=\frac{Au.Av}{\left \| Au \right \|.\left \| Av \right \|}\pm \mathcal{O}(\epsilon )
		\end{math}
	\end{center}
	This lemma provides a proof that the similarity between any two vectors can be preserved up to a factor of $\epsilon$ when these vectors are projected onto a random $t$-dimensional subspace. Such type of mapping can be performed by utilizing a matrix containing orthonormal columns as described in Lemma 5.2 \cite{kdd}. The lemma states that:\\
	Let $RP$ be a matrix of size $s \times t$ where $t<s$. Each of the entries of $RP$ are i.i.d. Gaussian random variable with zero mean and variance $\frac{1}{s}$, $r_{ij}\sim s(0, \frac{1}{s})$, $i$=1,$\cdots$ ,$s$ , $j$=1,$\cdots$, $t$.
	Let $W=RP^{T}RP$ and $W^{'}= RPRP^{T}$; then,
	
	\begin{align}
	E(w_{i,j})=\left\{\begin{matrix}
	1, & i=j \\ 
	0, & i \neq j
	\end{matrix}\right.
	\ \ \ \ \
	Var(w_{i,j})=\left\{\begin{matrix}
	\frac{2}{s}, & i=j \\ 
	\frac{1}{s}, & i \neq j
	\end{matrix}\right.
	\end{align}
	
	\begin{align}
	E(w_{i,j}^{'})=\left\{\begin{matrix}
	\frac{t}{s}, & i=j \\ 
	0, & i \neq j
	\end{matrix}\right.
	\ \ \ \ \
	Var(w_{i,j}^{'})=\left\{\begin{matrix}
	\frac{2t}{s^{2}}, & i=j \\ 
	\frac{t}{s^{2}}, & i \neq j
	\end{matrix}\right.
	\end{align}
	
	where, $w_{i,j}$ and $w_{i,j}^{'}$ are elements of $W$ and $W^{'}$, respectively.
	
	The output here confirms that $E[RP^{T}RP]=I$, where $I$ denotes an identity matrix. The elements of $RP^{T}RP$ are centered around their mean with very small variance. This suggests that vectors with random directions are close to orthogonal (i.e. $RP^{T}RP \approx I$). Further, it is obvious that if $r_{ij} \sim s(0, \frac{1}{s})$, then, $E[\left \| r_{j} \right \|^{2}]=E[\sum_{i=1}^{s}r_{ij}^{2}]=1 \  \text{and} \ Var[\left \| r_{j} \right \|^{2}]=Var[\sum_{i=1}^{s}r_{ij}^{2}]=\frac{2}{s}$, where $r_{j}$ denotes individual columns of $RP$. This mathematical proof ensures that columns in $RP$ are saturated around one which signifies that the vectors in $RP$ are nearly orthonormal. For revocable biometric template generation, we evaluate the probability of false match when the biometric data of the same user is exploited with the different random projection matrices, denoted as $P_{fm}$. Therefore, the revocability i.e. probability of a protected template being revocable can be defined as: $P_{r}=1-P_{fm}$. The higher value of $P_{r}$ corresponds to better revocability. In general, zero $P_{fm}$ cannot be obtained if we apply random projection directly onto the biometric data ($LT$). Further, this probability can be reduced by adding an extra vector $d \in \Re^{s}, d_{i}>> th$ to the $LT$, ${LT}'=LT+d$, where $th$ denotes the threshold of verification system \cite{rproj}. In similar manner, the biometric templates with $P_{r} \approx $ 1 could be derived, if different random projection matrices are exploited on the original template of the same user nullifying the record multiplicity attack \cite{crack}. In this work, we achieve $P_{r}$= 0.982 corresponding to a threshold, $th$=0.65.
	
	We also verify this security aspect empirically by generating 100 different protected templates using 100 different projection matrices from the same finger. Next, we perform a comparison of these 100 templates with the originally enrolled template to obtain the pseudo-imposter scores. We achieve mean and variance of (0.7519; 0.018), (0.3982; 0.177) and (0.3563; 0.189) for genuine, imposter and pseudo-imposter distributions, respectively. These values indicate that mean and variance for the pseudo-imposter distribution are at a distant to genuine distribution and near towards the imposter distribution. Moreover, we obtain FMR = 0, which depicts that all queries are rejected. This signifies that the derived templates are dissimilar to the enrolled templates for the same finger. Although, the templates are generated from the same finger pattern, they are uncorrelated to each other. Therefore, the claim of revocability is preserved.
	
	\subsection{Diversity}
	The characteristics of diversity state that it should derive numerous templates and these derived templates should not provide positive biometric claim over other applications to avoid cross-matching. In our method, multiple fingerprint templates can be derived by choosing the different projection matrices ($RP$) with the different seed values ($\kappa$). Also, the two parameters illustrated in Section IV(C); the number of sectors ($s$) and log-base value ($b$) can be utilized to derive numerous templates. The derived protected templates are sufficiently different from the raw fingerprint template which indicates that a user can enroll itself with different templates in different applications without any cross-matching. Hence, it has been confirmed that the method validates the property of diversity.
	
	\subsection{Other attacks}
	We also analyze the possibility of different types of attacks namely Attacks via Record Multiplicity, pre-image, cross-matching, distinguishing and annealing attacks to validate the robustness of the proposed work:

	\textit{Attack via Record Multiplicity (ARM)}: This is a scenario where the attacker employs multiple stolen protected templates with or without associated parameters to generate original template \cite{arm}. For our approach, it is infeasible to apply ARM since stored template's dimensionality is uncorrelated to the original template's feature space. Further, numerical values in protected template contain both positive and negative signs due to random projection based transformation. ARM attacks can be launched where the multiplication output contains either all positive or all negative entries. Therefore, the possibility of ARM attack for the proposed method gets nullified. 
	
	\textit{Pre-image attack}: In this attack, the attacker can utilize multiple protected instances to derive a pre-image instance. Knowledge of Security can also be challenged using feature order with different projection matrices to create a fake template. Biohashing based methods \cite{biohash,hash,hash2,palmhash} derive binary string by projecting feature vectors with user-specific random numbers. In contrast, the bit-string could be easily exploited to disclose original minutiae information. Moreover, the projection matrix in Biohashing is not only a square matrix but also have orthonormal row vectors i.e. $R_{proj} \cdot R_{proj}^{T}$=I, where $ R_{proj}^{T}$ is the pseudo-inverse of $R_{proj} $ and I is the identity matrix. This makes the Biohashing methods vulnerable to pre-image attack. However, the proposed random projection based transformation is different to the methods involving Biohashing. Here, the random projection is utilized to hide the log template among infinitely many possible solutions. Also, our method does not depend on the order of feature components while generating the original as well as the protected template. Further, any value could not be investigated from the two projected feature vectors in any position due to the different sized enrolled and query template. Hence, pre-image attack could not be utilized to derive the original template in our method.

	\textit{Cross-matching attack}: The cross-matching attack refers to the scenario where an adversary is able to compromise the databases stored in different applications.  The protected templates each from different applications are analyzed to restore the original template. However, the random projection transformation described in Eq. (6) avoids any possibility of cross-matching attack across different applications. The proof for the same has been stated in proposition 1 and 2 of Section III-E.

	\textit{Distinguishing attack}: In distinguishing attack \cite{dquad}, an imposter tries to utilize the same protected template captured from the different applications to derive the original template by correlating the information. To prevent this, the different protected templates can be utilized in the different applications. However, the attacker can retrieve the different protected templates ($CT$) along with the known random projection matrices ($RP$) from the different applications in the known-key distinguishing attack. In this situation, the attacker would be able to unveil log template ($LT$). Further, he may estimate the paired output ($CP$) or approximate $CP$ using the value of $b$ or approximately equal to $b$. However, it is not possible to derive original ridge features since the Cantor pairing function is irreversible as defined in Eq. (12a)-(12d) (for details, see Section \ref{nonin}).
	
	\textit{Annealing attack}: In this attack \cite{anneal}, the protected template is divided into multiple regions, and some regions of a sample template are paired with some regions of the reference template to evaluate similarity score. If the similarity score exceeds the threshold, the vicinity corresponding to sample's region is included in the gummy template. This step is repeated until it outputs a gummy template including all matched vicinities. Our approach is robust against this type of attack due to the following reasons:
	\begin{enumerate}
		\item Our approach evaluates the nearest neighbor minutia for each minutiae point causing the different radii to the different minutiae points. Hence, it is very hard to map the gummy template with the original template which is derived from the multiple regions with the variable radius.
		\item Ridge-based features are utilized for the neighboring minutiae in each sector instead of relative distances or the directional difference between minutiae pairs. Here, the measured ridge features are invariant to the inter-ridge distances and locations of minutiae points. 
	\end{enumerate}

\section{Conclusion and future work}
In this paper, we have proposed pre-alignment free cancelable fingerprint template generation technique. The proposed technique does not rely on detection of singular points. We divide the input fingerprint image into a number of sectors of equal angular width considering each minutia as a reference and use the nearest neighbor minutiae in each sector to compute transformation invariant ridge count and mean ridge orientation features from each sector. Cantor pairing function is applied to encode these features uniquely. Further, the pointwise logarithm operation is exploited to yield uniformly distributed features. Finally, a random projection is adopted to derive a non-invertible and revocable cancelable template. Experimental evaluation performed over four datasets of FVC2002, FVC2004, and FVC2006 databases depict that the significant performance improvement is achieved as compared to the current state-of-the-art techniques. Moreover, the security analysis of our work confirms that our approach fulfills the desired characteristics of template protection schemes preserving the recognition performance too. However, the proposed ridge-based feature computation for low-quality fingerprint and partial fingerprint images is still a demanding area. In our future work, we would try to address this limitation. The computational complexity of the proposed method is $\mathcal{O}(n^{2})$. Hence, future work along the direction to reduce/improve computational complexity is underway. 
	
%
	
		\bibliography{refs}
	
%

	\EOD
	
\end{document}